\documentclass[final]{cvpr}

\usepackage{times}
\usepackage{epsfig}
\usepackage{graphicx}
\usepackage{amsmath}
\usepackage{float}
\usepackage{amssymb}
\usepackage{caption}
\usepackage{balance}
\usepackage{subcaption}

\usepackage{makecell}
\usepackage{multirow}
\usepackage[table,xcdraw]{xcolor} 
\usepackage{tabularx}
\usepackage[font=small,skip=3pt]{caption}

\usepackage[pagebackref=true,breaklinks=true,colorlinks,bookmarks=false]{hyperref}

\def\cvprPaperID{5521} 
\def\confYear{CVPR 2021}

\newcommand{\std}[1]{{\footnotesize{$\pm$#1}}}
\newcolumntype{P}[1]{>{\centering\arraybackslash}m{#1}}
\usepackage{soul} 

\begin{document}

\title{Spoken Moments: Learning Joint Audio-Visual Representations from Video Descriptions}

\author{Mathew Monfort\thanks{equal contribution}\\
MIT\\
{\tt\small mmonfort@mit.edu}
\and
SouYoung Jin\footnotemark[1]\\
MIT\\
{\tt\small souyoung@mit.edu}
\and
Alexander Liu\\ 
MIT\\
{\tt\small alexhliu@mit.edu}
\and
David Harwath\\
UT Austin\\
{\tt\small harwath@cs.utexas.edu}
\and
Rogerio Feris\\
IBM Research\\
{\tt\small rsferis@us.ibm.com}
\and
James Glass\\
MIT\\
{\tt\small glass@csail.mit.edu}
\and
Aude Oliva\\
MIT\\
{\tt\small oliva@mit.edu}
}

\maketitle

\begin{abstract}
    When people observe events, they are able to abstract key information and build concise summaries of what is happening. These summaries include contextual and semantic information describing the important high-level details (what, where, who and how) of the observed event and exclude background information that is deemed unimportant to the observer. 
    With this in mind, the descriptions people generate for videos of different dynamic events can greatly improve our understanding of the key information of interest in each video. 
    These descriptions can be captured in captions that provide expanded attributes for video labeling (e.g. actions/objects/scenes/sentiment/etc.) while allowing us to gain new insight into what people find important or necessary to summarize specific events. 
    Existing caption datasets for video understanding are either small in scale or restricted to a specific domain.  To address this, we present the Spoken Moments (S-MiT) dataset of 500k spoken captions each attributed to a unique short video depicting a broad range of different events.  We collect our descriptions using audio recordings to ensure that they remain as natural and concise as possible while allowing us to scale the size of a large classification dataset.
    In order to utilize our proposed dataset, we present a novel Adaptive Mean Margin (AMM) approach to contrastive learning 
    and evaluate our models on video/caption retrieval on multiple datasets.  We show that our AMM approach consistently improves our results and that models trained on our Spoken Moments dataset generalize better than those trained on other video-caption datasets.
    
    \small{\url{http://moments.csail.mit.edu/spoken.html}}
\end{abstract}
\section{Introduction}


Video understanding has typically been focused on action recognition and object tracking as the temporal aspect of videos lends itself strongly to the task of representing motion, a key component of an action. 
Breaking down video analysis to simple tasks, such as action recognition, allows for efficient data annotation for building large datasets to train deep learning models \cite{kay2017kinetics,monfortmoments,goyal2017something} which has been extremely successful for images with object annotations \cite{krizhevsky2012imagenet}.
A main difficulty is that, in contrast to an image, a video often captures an interaction between agents and objects that evolves over time.  These interactions can be as simple as ``a person picking up a glass of water", but even in this case three different objects (``person", ``glass" and ``water") are included in the interaction. 
Additionally, the video may also continue to depict the ``person drinking from a glass" and the ``person putting the glass back down on the table". These sequential events present additional challenges for video datasets where single annotations may not be sufficient to explain the events depicted.  Multi-label approaches to video annotation have attempted to address this problem by labeling multiple actions in a video \cite{monfort2019multimoments,Gu_2018_CVPR,zhang2018multi}.  However, these methods focus on single domain annotations, such as actions or objects, and do not capture additional contextual information, such as ``person \emph{angrily} putting down the \emph{dirty} glass on a \emph{rusted} table",  which can change the interpretation of an event and how it fits into a sequence of observations.

A solution for capturing more fully the content of video is to annotate multiple actions or objects in each video \cite{Gu_2018_CVPR,yeung2015every,monfort2019multimoments,real2017youtubeboundingboxes}. However labels like ``drinking", ``glass", only provide a portion of the information needed to interpret the veracity of the event. 
Additional narratives may include intuitive descriptions and intentions, such as ``an exhausted man picks up a dirty glass of water and drinks from it before angrily putting it down on a table" which would dramatically change the event interpretation. The full lingual description combines these actions with adjectives and nouns (objects) that contextualize the events depicted leading to a better understanding of the video.  This is our goal in providing a new large scale dataset for training models for full video understanding.

        
We introduce a large scale video caption dataset, Spoken Moments in Time (S-MiT), to allow large deep learning models for video understanding to learn contextual information. 
Most existing video description datasets \cite{xu2016msr-vtt, Sigurdsson2016HollywoodIH, krishna2017dense, gella-etal-2018-dataset,youcook2} are limited in size when compared to the large datasets for action recognition \cite{kay2017kinetics,monfortmoments,goyal2017something}.  A likely cause is the increased cost of collecting full text descriptions for videos compared to single label annotations.  Recent work in image captioning \cite{david2020ijcv} addressed this problem by collecting audio descriptions for a large set of images from the Places dataset \cite{zhouKLTO16}.  
Collecting spoken captions is faster and more efficient due to the low overhead of speaking compared to typing. In addition, recording of spontaneous speech rather than typed text can produce more natural descriptions of an event.
An automatic speech recognition (ASR) system was then used to transcribe the spoken descriptions to text captions. In this work, both audio, text and video models were jointly trained 
via contrastive learning to learn joint cross-modal representations. 
We build on this approach and compare models that learn directly from the spoken captions to models that include a trained ASR model which feeds generated text transcriptions into an NLP language model. We then jointly train caption and visual models (based on concatenated video and image features) using a novel Adaptive Mean Margin (AMM) approach to contrastive learning to align the visual and caption representations.
We evaluate our models on multiple datasets for video/caption retrieval and show that a model trained using AMM on S-MiT achieves the best general performance across 
four datasets.

\begin{figure*}[t]
\centering
    \scalebox{0.98}{
    \includegraphics[width=\linewidth, trim={0 1cm 0 0},clip]{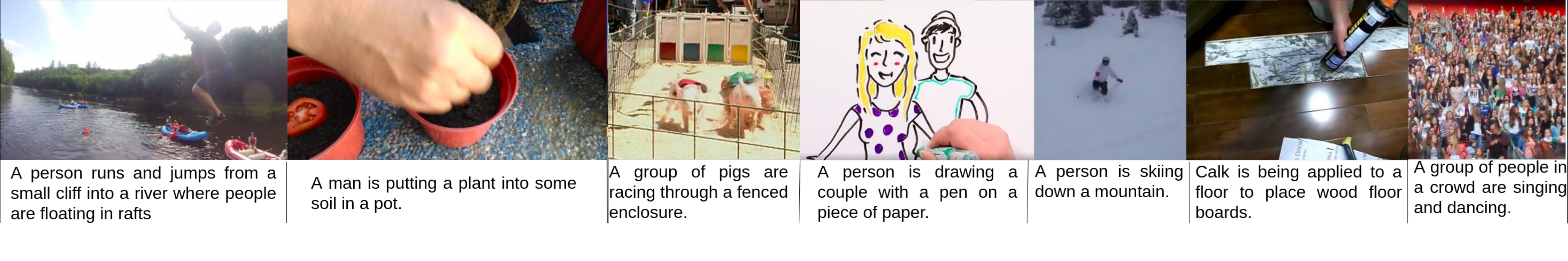}
    }
    \caption{\textbf{Examples from the Spoken Moments Dataset:} The dataset is composed of videos and the corresponding spoken captions. We show some examples of the text transcriptions, automatically generated using the public Google ASR engine.}
    \label{fig:spoken_examples}
\end{figure*}

Altogether, our novel contributions include:

\begin{enumerate} 
    \item  The large-scale \textbf{Spoken Moments in Time dataset} (S-MiT) which includes 500k pairs of video clips and corresponding audio descriptions.  This new dataset represents the largest video description dataset available and will serve as a new benchmark for the community.
    \item \textbf{Benchmark models} with aligned spoken caption and video representations learned via contrastive learning.  We compare approaches that learn directly from the spoken descriptions as well as approaches that include ASR transcriptions that feed into different language models to generate caption representations.
    \item An \textbf{Adaptive Mean Margin} (AMM) approach to cross-entropy based contrastive  learning.
\end{enumerate}
\section{Related work}
\subsection{Video Understanding}

The field of video understanding has recently seen fast progress partly due to the availability of large scale video datasets including ActivityNet \cite{caba2015activitynet}, Kinetics \cite{kay2017kinetics},  Moments in Time \cite{monfortmoments,monfort2019multimoments} and YouTube-8M \cite{youtube8m}.  These large datasets are used to pretrain models that are fine-tuned on smaller action recognition datasets such as UCF101 \cite{soomro2012ucf101} and HMDB \cite{10.1007/978-3-642-33374-3_41}. 
With the increased availability of large scale video datasets, many different models have been proposed to improve performance on a number of video understanding tasks. Two-stream convolutional neural networks (CNNs) combine optical flow with RGB frames to capture both temporal and spatial information \cite{simonyan2014two}. 
I3D models \cite{carreira2017quo} combine 3D CNNs \cite{tran2015learning}, which use a 3D kernel to learn temporal information from a frame sequence, with optical flow to form a two-stream 3D network ``inflated'' from 2D filters pre-trained on ImageNet \cite{deng2009imagenet}.  More recently a temporal shift module has been proposed to integrate temporal information into 2D models by shifting frame representations across the temporal dimension \cite{Lin_2019_ICCV}.

Recently multi-modal visual understanding methods have received significant attention~\cite{david2020ijcv,Suris_2019_CVPR,merkx2019interspeech,vasudevan2018wacv,ilharco2019largescale,groundedWordsVideo}. The DAVEnet model \cite{david2020ijcv} has been proposed for jointly learning aligned representations between images and spoken captions, and has been extended to align frame-wise video representations with synchronized audio narration for cross-modal audio-visual concept learning \cite{groundedWordsVideo}. Here, we build on the motivation from this paper and \textbf{learn aligned representations between videos and unsynchronized spoken descriptions} using the S-MiT Dataset.

\subsection{Caption Datasets}

There have been a number of different datasets released for providing language descriptions of visual information.  Flickr8k \cite{Hodosh2013FramingID} and Flickr30k \cite{Plummer_2015_ICCV} include 8k and 30k images respectively each sourced from Flickr.  Each image is associated with 5 text captions describing what is in the image.  An additional set of 5 audio captions per image in both sets was recently collected for learning joint embeddings between speech and images \cite{david2020ijcv}.  The Visual Genome dataset \cite{krishnavisualgenome} includes captions for multiple regions of more than 180k images allowing for fine-grained descriptions of each image.  The Places Audio Caption dataset \cite{david2016nips} contains approximately 400k images from the Places 205 \cite{NIPS2014_5349} image dataset with audio captions of people verbally describing each image. MS COCO \cite{capeval2015} is a large image dataset for object recognition, segmentation, and captioning which includes roughly 1 million captions for 160k Flickr images.  Conceptual Captions \cite{sharma2018conceptual} contains 3.3M images with captions generated from HTML attributes associated with web based images. The Stock3M dataset \cite{Wang_2017_CVPR} includes 3.2 million images each with a crowdsourced caption.

Beyond the numerous datasets available or image captioning \cite{Hodosh2013FramingID,Plummer_2015_ICCV,krishnavisualgenome,capeval2015,sharma2018conceptual,Wang_2017_CVPR}, including those that provide spoken descriptions \cite{david2016nips,david2020ijcv}, there are a variety of different video caption datasets available. A number of these datasets are related to cooking \cite{tacos:regnerietal:tacl,cispa1826,10.1007/s11263-015-0851-8,Damen2018EPICKITCHENS,Damen2020RESCALING} including YouCook \cite{DaXuDoCVPR2013} and YouCook II \cite{youcook2} which include 2k videos from YouTube each with multiple captions annotated at different segments of each video. MPII-Movie Description Corpus \cite{7298940} contains transcribed audio descriptions from 94 Hollywood movies split into 68k clips 
where each clip is paired with a sentence from the movie scripts and an audio description of the visual content in each clip.  Similarly, 
Large Scale Movie Description Challenge (LSMDC) dataset \cite{lsmdc} contains 200 movies with 120K sentence descriptions. 
VideoStory \cite{gella-etal-2018-dataset} contains 20k social media videos where each video contains a paragraph length description. The ActivityNet Captions dataset \cite{krishna2017dense} has 20k videos with 100k text descriptions. 
The Microsoft Video Description (MSVD) dataset \cite{chen-dolan-2011-collecting} contains 2k YouTube clips with a 10-25 second duration and an average of 41 single sentence descriptions per clip.  
MSR-Video to Text (MSR-VTT) \cite{xu2016msr-vtt} contains 7k videos split into 10k clips with 20 captions per video. 

HowTo100M \cite{miech19howto100m} contains 136 million clips sourced from 1.22 million instructional videos with narrations generated from subtitles associated with each video. However, the subtitles are not human verified captions and the content is constrained to instructional videos.  Since the text associated with the clips in the HowTo100M dataset are transcriptions of a narrator completing a task in the video, the short text phrases from the subtitles occasionally share noisy associations with the reference clip. In Section \ref{sec:results}, and Table \ref{table:caption_stats}, we decided to compare our contributions using strict caption datasets as we are proposing a large-scale human annotated caption dataset with full human generated descriptions for each video.

VaTeX \cite{Wang_2019_ICCV} contains 41k videos sourced from the Kinetics-600 dataset \cite{kay2017kinetics,carreira2018short} annotated with 10 English captions and 10 Chinese captions for multilingual captioning.  VaTeX is the most similar to our proposed dataset in that it is sourced from an existing video dataset for action recognition and the captions are directly annotated. 

In this work, we present a new dataset, \emph{Spoken Moments in Time (S-MiT)}, which includes spoken audio captions for 500k unique three second clips each with different source videos from the Moments in Time dataset \cite{monfortmoments,monfort2019multimoments}.  In addition to vast increase in scale over other video-caption datasets, a major contribution is that we are using spoken descriptions rather than text. This allows us to train spoken caption models to directly align with video models. This is not possible with the other large video caption datasets and allows for spoken caption models to be analyzed with matching video information. We also show that models trained on our S-MiT dataset generalize much better in retrieval to the video-caption pairs in other datasets. This is due to the large coverage, diversity and scale of our proposed dataset.

\subsection{Cross Modal Contrastive Learning}
Cross modal learning has been used to jointly self-supervise audio-visual models \cite{NIPS2016_6146, owens2016, Zhao_2018_ECCV} with synchronized information while NLP approaches have been leveraged to align joint representations for both visual and language modalities using spoken and text descriptions \cite{alayrac:hal-01171193,  ZhLoCoBMVC18}. This is typically done via Contrastive Learning where the alignment between positive pairs (language and visual input) is trained to be stronger than those of non-positive pairs \cite{1640964}.
For visual representations, a triplet based max-margin loss is commonly used to discriminate representations between positive and negative pairs \cite{zhang2016colorful, 8099559, 7410524}. Semi-hard negative mining \cite{Schroff_2015_CVPR} and a dot-product based similarity score have been used to jointly learn audio-visual embeddings between images and spoken captions \cite{david2020ijcv} while batch-wise cross-entropy approaches to contrastive learning have been used to increase the amount of information utilized in learning by considering all negative examples in a mini-batch \cite{DBLP:journals/corr/abs-1807-03748, chen2020simple}. Work on bidirectional speech/image retrieval using audio descriptions of images integrated ideas from max-margin contrastive learning and added a margin into the cross-entropy loss \cite{ilharco2019largescale}. SimCLR \cite{chen2020simple} added a non-linear projection head that maps paired representations into a common space allowing for stronger representations.

A pretrained language model has recently been used to improve cross-modality learning with language and visual input pairs.
VilBERT \cite{NIPS2019_8297} added a pretrained BERT \cite{DBLP:conf/naacl/DevlinCLT19} transformer to capture semantic language representations associated with object detection proposals from a pretrained faster RCNN network. VideoBERT~\cite{videoBert2019iccv} extended BERT to jointly learn the visual and linguistic domain by generating tokenized visual words. 
Inspired by this prior work, we propose adding a \textbf{pretrained language model that maps word predictions from a trained ASR model to semantic language features} in order to generate rich spoken caption representations.  We then utilize an MLP to project these caption representations, and our video representations, to an aligned joint representation which can be used for video/caption retrieval (see Section \ref{sec:results}).

\subsubsection{Optimization Approaches}
\label{sec:related_loss}
A common approach to optimization in contrastive learning settings is to use a similarity based loss function. We formulate the contrastive loss as, 
    $\mathcal{L}=\mathcal{L}_{vc}+\mathcal{L}_{cv}$,
where the goal is to maximize the discrimination between positive and negative paired captions $c$ and videos $v$.  The loss is split into two tasks where $\mathcal{L}_{vc}$ forms pairs from a fixed video and each caption in a sampled mini-batch, while $\mathcal{L}_{cv}$ fixes the caption and forms pairs with each video in the mini-batch. Below we discuss different approaches of $\mathcal{L}_{xy}$, where $x$ and $y$ are interchangeable with $v$ and $c$.


Semi-hard negative mining (SHN) \cite{Schroff_2015_CVPR} has been used for learning aligned cross-modal embeddings using a triplet loss \cite{david2020ijcv,8461684}.  This is an improvement over hard negative mining \cite{DBLP:journals/corr/FaghriFKF17} since a sampled negative example is constrained to be less similar to the anchor than the positive sample while still being within the margin and thus contributing a loss at each step with the margin $M=1$, 
$\mathcal{L}_{xy} = \max(\mathcal{S}(x_i,y_j) - \mathcal{S}(x_i,y_i) + M, 0)$,
where $\mathcal{S}(x_i,y_j)$ is a similarity score for the representations of $x_i$ and $y_j$, with $x_i$ and $y_i$ forming a positive pair.  


Noise contrastive estimation (NCE) \cite{pmlr-v9-gutmann10a} has been applied to 
contrastive learning \cite{chen2020simple, DBLP:journals/corr/abs-1807-03748} by using a log-likelihood based loss function that learns to discriminate between positive and negative pairs of feature embeddings,
\begin{align}
    \text{\footnotesize $\mathcal{L}_{xy} = -\frac{1}{B}\sum^{B}_{i=1}log\frac{e^{\mathcal{S}(x_i,y_i)}}{\sum^{B}_{j=1}\mathit{I}_{i \neq j}e^{\mathcal{S}(x_i,y_j)}}$},
    \label{eq:nce}
\end{align}
where $\mathit{I}_{i \neq j}$ is an indicator function that we only considers negative pairs in the denominator. This has been shown to improve feature alignment compared to SHN \cite{chen2020simple}.


Masked Margin Softmax Loss (MMS) \cite{ilharco2019largescale} and Large Margin Cosine Loss (LMCL) \cite{Wang_2018_CVPR} incorporate a positive margin into the contrastive learning framework in order to improve feature discrimination among non-paired embeddings.  MMS uses a monotonically increasing margin to allow for initial learning to begin to converge before a large alteration to the loss is added. LMCL proposes a theoretical limit on the maximum margin size of $1-cos\frac{2\pi}{N}$ where $N$ refers to the number of classes being discriminated.  For aligning captions to visual information, the class size can be considered unbounded as each caption represents a slightly different representation that we want to discriminate leading to a max margin size of $1$. Concretely, MMS proposes adding a margin to Equation \ref{eq:nce}, 
\begin{align}
    \text{\footnotesize $\mathcal{L}_{xy} = -\frac{1}{B}\sum^{B}_{i=1}log\frac{e^{\mathcal{S}(x_i,y_i)-M}}{e^{\mathcal{S}(x_i,y_i)-M}+\sum^{B}_{j=1}\mathit{I}_{i \neq j}e^{\mathcal{S}(x_i,y_j)}}$},
    \label{eq:margin_nce}
\end{align}
where the margin, $M$, starts as 0.001 and is exponentially increased by a factor of 1.002 every 1000 training steps.

We propose extending the idea of an increasing margin in MMS to an adaptive setting that does not require setting the initial value of the margin or the growth rate.  We refer to this approach as an Adaptive Mean Margin (AMM) where the margin is set as the mean distance between the positive pair and the set of negative pairs in a batch.  We describe AMM in more detail in Section \ref{sec:approach}.

\section{The Spoken Moments Dataset}
We begin with the Moments in Time dataset \cite{monfortmoments} as it includes over 1 million videos 
sourced 
from a number of different video hosting sites with strong inter \& intra-varietal variation in terms of the number of events depicted in each video. Further, the videos are all cut to 3 seconds allowing for a concise description to effectively capture the localized information of each event.  Here we refer to concise descriptions as those that focus on key events depicted in the video and does not imply partial descriptions. In data collection, annotators may watch a video as many times as desired. During recording, we block the annotators from seeing/hearing the video to encourage descriptions of important memorable events rather than every specific detail. This approach does not preclude the annotators from describing sequential or simultaneous events as shown in our qualitative examples (see Figure \ref{fig:spoken_examples}).  We describe our annotation approach in more detail in the supplementary material.
    
\subsection{Dataset Statistics}


\begin{figure*}
\begin{minipage}{0.39\textwidth}
\setlength\tabcolsep{3 pt}
    \centering
        \scalebox{0.6}{
        \begin{tabular}{c c c c}
        \Xhline{0.8pt}
        Type & Total & Average & Unique \\
        \Xhline{0.8pt}
        Words & 5,618,064 & 18.01 & 50,570 \\
        Verbs & 492,941 & 1.58 & 7,436 \\
        Nouns & 1,365,305 & 4.37 & 20,645 \\
        Adjectives & 386,039 & 1.24 & 12,523 \\
        \Xhline{0.8pt}
        \end{tabular}}
        
\vspace{5 pt}

\setlength\tabcolsep{3 pt}
    \centering
        \scalebox{0.6}{
        \begin{tabular}{ c | c c }
        \Xhline{0.8pt}
        Type & Dataset & Coverage \\
        \Xhline{0.8pt}
        \multirow{2}{*}{Objects} & ImageNet & 69.2\% \\
        & MS-COCO & 100\% \\
        \hline
        \multirow{2}{*}{Actions} & Kinetics  & 85.1\% \\
        & Moments in Time & 96.2\% \\
        \hline
        Scenes & Places365 & 47.4\% \\
        \Xhline{0.8pt}
        \end{tabular}}
    \label{table:coverage}
\end{minipage}
\begin{minipage}{0.59\textwidth}
\setlength\tabcolsep{3 pt}
    \centering
    \scalebox{0.6}{
    \begin{tabular}{ c | c c c c c c c}
        \Xhline{0.8pt}
        \textbf{Dataset} & \textbf{Clips} & \textbf{Videos} & \textbf{Captions} & \textbf{Words} & \textbf{Vocab} & \textbf{Domain} & \textbf{Spoken}\\
        \Xhline{0.8pt}
        TACoS \cite{tacos:regnerietal:tacl} & 7,206 & 127 & 18,227 & 146,771 & 28,292 & Cooking &\\
        YouCook II \cite{youcook2} & 15,400 & 2,000 & 15,400  & 121,418 & 2,583 & Cooking &\\
        MSVD \cite{chen-dolan-2011-collecting} & 1,970 & 1,970 & 70,028 & 607,339 & 13,010 & General & \\
        Charades \cite{Sigurdsson2016HollywoodIH} & 10,000 & 10,000 & 27,800 & 645,636 & 32,804 & General & \\
        MPII-MD \cite{7298940} & 68,337 & 94 & 68,375 & 653,467 & 24,549 & General & \\
        MSR-VTT \cite{xu2016msr-vtt} & 10,000 & 7,180 & 200,000 & 1,856,523 & 29,316 & General & \\
        ActivityNet Captions \cite{krishna2017dense} & 100,000 & 20,000 & 100,000 & 1,348,000 & 15,564 & General & \\
        VideoStory \cite{gella-etal-2018-dataset} & 123,000 & 20,000 & 123,000 & 1,633,226 & - & General &\\
        Epic-Kitchens \cite{Damen2018EPICKITCHENS,Damen2020RESCALING} & 76,885 & 633 & 76,885 & 227,974 & 1,737 & Cooking &\\
        Vatex-en \cite{Wang_2019_ICCV} & 41,300 & 41,300 & 413,000 & 4,994,768 & 44,103 & General & \\
        \textbf{Spoken Moments} & \textbf{515,912} & \textbf{459,742} & \textbf{515,912} & \textbf{5,618,064} & \textbf{50,570} & \textbf{General} & \textbf{\checkmark}\\
        \Xhline{0.8pt}
    \end{tabular}}
\end{minipage}
\caption{\textbf{Dataset Statistics:} On the top-left we show the total and average number of words, verbs, nouns and adjectives in our captions as well as the number of unique examples of each.  On the bottom-left we show the percentage of the class vocabulary from different datasets that occur in our captions.  On the right we compare our proposed Spoken Moments dataset to existing video caption datasets.  The word count and vocabulary for S-MiT are generated using ASR transcriptions.}
\label{table:caption_stats}
\end{figure*}

Our proposed Spoken Moments dataset contains 500k videos randomly chosen from the Multi-Moments in Time (M-MiT) training set and all of the 10k videos from the validation set.  Each video in the training set contains at least one audio description. 
We transcribed each audio recording using the public Google Automatic Speech Recognition (ASR) engine to generate text captions for each video.  When analyzing these transcriptions, we build a picture of the coverage and diversity of our captions.  Table \ref{table:caption_stats} (left) shows that our captions have an average length of 18 words with a unique vocabulary of 50,570 words consisting of 20,645 nouns, 12,523 adjectives and 7,436 verbs with a total word count of 5.6 million.  Table \ref{table:caption_stats} (right) shows a comparison of our Spoken Moments dataset to other existing datasets for video captioning.  Our dataset will be the largest public dataset in terms of video clips, source videos, total number of captions, total words in the captions and the vocabulary set of unique words occurring in the captions.  The increase in vocabulary size is important as it shows that our increase in the number of videos over previous datasets does not simply include repeated events but covers a novel breadth of information.  We can see the opposite effect of this in YouCook II \cite{youcook2} where the restricted domain of cooking videos results in a limited vocabulary used in the descriptions.

To understand how this vocabulary covers the class labels typically used for training computer vision models, we examined whether these labels exist in our vocabulary.  Table \ref{table:caption_stats} (right) shows that we have strong coverage of the two largest action recognition datasets for video understanding (Kinetics \cite{kay2017kinetics} and M-MiT \cite{monfort2019multimoments}).  We expected a large coverage of the events in M-MiT as we sourced our videos from this dataset and the action labels themselves are fairly general (e.g. ``running" and ``cooking").  For Kinetics, the labels are commonly tied to a noun preceded by a verb (e.g. ``brushing hair").  For these labels we consider them to exist in our dataset if both the verb and noun are in the same caption.  For example, ``A boy is in a bathroom brushing his teeth" would cover the class ``brushing teeth".  With this approach we see a 85.1\% coverage of the classes in Kinetics and a 96.2\% coverage of the classes in M-MiT showing a strong level of event diversity.  Similarly we see a strong overlap of the object classes in MS-COCO \cite{lin2014microsoft} (100\%) and ImageNet \cite{deng2009imagenet} (69.2\%) in our captions.  ImageNet coverage is likely lower due to the specific labels used for many of its classes (e.g. ``coucal").  Still, 69.2\% coverage means 692 ImageNet classes appear in our captions.  Similarly, Places \cite{NIPS2014_5349} scene labels are very specific and don't necessarily match the language used in our descriptions. For example, an ``abbey" will typically be described as a ``church" or ``monastery" in our captions.  We did not account for all of the synonyms possible 
and are only considering direct matches in our captions.  Even so we are able to find a 47.4\% coverage of the scene labels in Places365 in our dataset.

Here we provide information on some additional characteristics of our data that may be of interest.   While we do not release demographic info of our annotators or captions, about 57\% of the spoken captions were recorded by male voices and 43\% female.  For the audio streams of the videos, roughly 51\% include natural sound, 5\% have music as the audio and 44\% have no audio. This is consistent with the M-MiT dataset~\cite{monfort2019multimoments} from which we source our videos. Additionally, we found that less than 3\% of the videos contain captions that describe non-visible events (e.g. a car horn when no car is visible in the video frames).  For this reason we have chosen to focus our approach on learning a strong visual model in Section \ref{sec:model}.
\section{Learning Audio-Visual Representations}
\label{sec:model}
\begin{figure}
    \centering
    \includegraphics[width=0.9\linewidth]{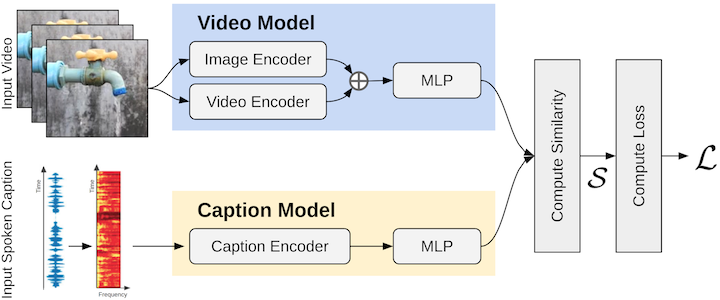}
    \caption{\textbf{Architecture:}
    Videos and captions are fed into the video/caption models where the outputs are used to compute a similarity matrix, $\mathcal{S}$, which is used to compute a loss, $\mathcal{L}$.}
    \label{fig:model_architecture}
\end{figure}
    
In order to learn from the large set of spoken captions in the proposed S-MiT dataset, we adopt a cross-modal architecture used in prior work \cite{miech19howto100m,david2020ijcv,AVLnet} which is composed of a video model and a caption model as depicted in  Figure~\ref{fig:model_architecture}. Specifically, we take $N$ video-caption pairs as input and encode each modality into a $4096$-D feature vector. We do this by adding a multilayer perceptron (MLP) as a projection head on top of both the video and the caption model.  This projection head is composed of two linear layers followed by gated linear units (GLU)~\cite{GLU}. We then compute the dot product between the video and caption representations to produce an $N$x$N$ similarity matrix, $\mathcal{S}$, which is used to compute our contrastive loss for training. In Section~\ref{sec:approach}, we describe our modified approach to margined contrastive learning which uses an Adaptive Mean Margin (AMM) which automatically adjusts itself during learning to improve the optimization signal during training.

\subsection{Video Model}
 \label{sec:video_model}
 Following prior work ~\cite{miech19howto100m}, we use two encoders to represent input videos: image \& video encoders. Specifically, we use a ResNet-152~\cite{resNet} pretrained on ImageNet~\cite{krizhevsky2012imagenet} and a temporal shift module (TSM) ResNet-50 model~\cite{Lin_2019_ICCV} pretrained on M-MiT~\cite{monfort2019multimoments}. Each encoder outputs a $2048$-D feature vector after max-pooling over the temporal dimension (8 frames for the TSM ($\sim$3 fps) and 3 frames for the image model (1 fps)). We concatenate the two $2048$-D vectors and feed the concatenated vector into an MLP projection head to get the final $4096$-D visual representation.  We examine the effect of using the image and video encoders as well as different pretrained models in the supplementary material.



\subsection{Caption Model}
\label{sec:caption_model}

\subsubsection{Language Caption Model}
\label{sec:language_caption_model}
Prior work in learning joint representations between audio captions and visual models has shown that utilizing ASR transcriptions greatly improves results \cite{david2020ijcv}.  We build on this idea and use the predicted words from a pretrained ASR model (e.g. Google's public ASR engine) to train our models.  
Concretely, we examine the effect of using different pretrained language models stacked on top of the ASR model predictions.  We begin by comparing the results of using Fasttext \cite{bojanowski2016enriching}, BART \cite{lewis2019bart} and BERT \cite{DBLP:conf/naacl/DevlinCLT19} models to generate semantic and contextual word representations for our captions.  During training, we randomly select 10 words from each caption to be included in training.  In the case of the BART and BERT models,  this selection happens after the full transformer model has been applied to avoid altering the results from the self-attention mechanisms.  If less than 10 words occur in a caption then we allow words to be sampled multiple times in the random selection.  This training augmentation allows different words in each caption to be represented differently at different training iterations. We examine the effect of this approach in the supplementary material. In test, we use the full transcription as input into the language model.  We average the word representations from the output of the language model to generate a single representation for each caption which we align to the video representations described in the previous section.  

\subsubsection{Spoken Caption Model}
We also train caption models with raw spoken captions instead of the corresponding transcription. 
For each caption, we randomly sample 10 seconds of speech for training and compute the 40-dimensional log Mel spectrogram to serve as the input of spoken caption model.
The input is fed into a spoken caption model where we consider ResDavenet~\cite{david2020ijcv} (which is designed specifically for speech) and two ImageNet ResNet~\cite{resNet} models (ResNet-34, ResNet-50). 
For the ResNet models, we modify the first convolutional layer to take the $1$-channel input so that spectrogram can be processed. 
In addition, the wav2vec~\cite{wav2vec} model, which takes raw waveform 
as the input, is also involved in our experiments. 
Spoken captions are first fed into the pre-trained wav2vec model, which produces $512$-D vectors per 210 ms.
We then feed them into a learnable ResStack, taken from ResDavenet, to learn representations of spoken captions.

\begin{table*}[tb]
\begin{minipage}{\linewidth}
    \centering
    \scalebox{0.6}{
    \begin{tabular}{ c | c  c  c  c | c  c  c  c | c  c  c  c }
        \Xhline{0.8pt}
        \textbf{Language} & \multicolumn{4}{c|}{\textbf{Caption to Video}} & \multicolumn{4}{c|}{\textbf{Video to Caption}} & \multicolumn{4}{c}{\textbf{Mean}} \\
        \textbf{Caption Model} & R@1 & R@5 & R@10 & mAP & R@1 & R@5 & R@10 & mAP & R@1 & R@5 & R@10 & mAP \\
        \Xhline{0.8pt}
        Fasttext \cite{bojanowski2016enriching} & 17.1\std{0.8} & 44.0\std{0.6} & 57.2\std{0.5} & 30.2\std{0.5} & 24.1\std{0.5} & 49.9\std{0.6} & 61.8\std{1.3} & 36.6\std{0.3} & 20.6\std{0.5} & 46.9\std{0.6} & 59.5\std{0.8} & 33.4\std{0.4}\\
        BERT \cite{DBLP:conf/naacl/DevlinCLT19} & 25.9\std{0.6} & 55.5\std{1.2} & 67.0\std{1.1} & 39.7\std{0.7} & 33.3\std{1.4} & 62.1\std{1.0} & 72.0\std{0.6} & 46.5\std{1.2} & 29.6\std{0.8} & 58.8\std{1.0} & 69.5\std{0.8} & 43.1\std{0.8}\\
        BART \cite{lewis2019bart} & \textbf{33.1}\std{0.9} & \textbf{65.5}\std{1.5} & \textbf{76.6}\std{1.3} & \textbf{47.8}\std{1.1} & \textbf{43.8}\std{0.7} & \textbf{71.5}\std{1.2} & \textbf{80.9}\std{1.6} & \textbf{56.4}\std{0.7} & \textbf{38.4}\std{0.4} & \textbf{68.5}\std{1.3} & \textbf{78.7}\std{1.4} & \textbf{52.1}\std{0.8}\\
        \Xhline{0.8pt}
    \end{tabular}}
\caption{\textbf{Language Caption Model Comparison on Video/Caption Retrieval:} Here we compare the video/caption retrieval results on the test set of the Spoken Moments dataset using models trained with three different language models.}
\label{table:language_model_comparison_retrieval}
\vspace{5mm}
    \scalebox{0.6}{
    \begin{tabular}{ c | c | c  c  c  c | c  c  c  c | c  c  c  c }
        \Xhline{0.8pt}
        \multirow{2}{*}{\textbf{Dataset}} & \multirow{2}{*}{\textbf{Loss}} & \multicolumn{4}{c|}{\textbf{Caption to Video}} & \multicolumn{4}{c|}{\textbf{Video to Caption}} & \multicolumn{4}{c}{\textbf{Mean}} \\
         & & R@1 & R@5 & R@10 & mAP & R@1 & R@5 & R@10 & mAP & R@1 & R@5 & R@10 & mAP \\
        \Xhline{0.8pt}
        \multirow{5}{*}{Vatex \cite{Wang_2019_ICCV}}
        & NCE & 43.6\std{1.4} & 77.4\std{1.4} & 86.5\std{1.4} & 58.4\std{1.2} & 39.4\std{1.3} & 74.3\std{1.0} & 84.7\std{0.8} & 54.7\std{1.0} & 41.5\std{1.2} & 75.8\std{1.1} & 85.6\std{1.1} & 56.5\std{1.0}\\
        & SHN & 19.6\std{1.4} & 50.2\std{1.5} & 63.9\std{0.6} & 33.8\std{1.1} & 22.9\std{1.0} & 54.0\std{0.9} & 68.8\std{1.2} & 37.6\std{0.9} & 21.3\std{0.9} & 52.1\std{0.8} & 66.3\std{0.8} & 35.7\std{0.7}\\
        & MMS & 46.2\std{1.5} & 79.7\std{0.8} & 88.1\std{0.8} & 60.7\std{1.0} & 42.0\std{0.7} & \textbf{77.7}\std{0.7} & \textbf{86.8}\std{0.3} & 57.5\std{0.6} & 44.1\std{1.1} & 78.7\std{0.7} & 87.4\std{0.5} & 59.1\std{0.7}\\
        & AMM & \textbf{48.7}\std{1.4} & \textbf{82.0}\std{0.9} & \textbf{89.3}\std{1.1} & \textbf{63.0}\std{1.0} & \textbf{43.0}\std{0.7} & 77.4\std{1.1} & 85.8\std{0.7} & \textbf{58.3}\std{0.6} & \textbf{45.9}\std{1.0} & \textbf{79.7}\std{0.4} & \textbf{87.5}\std{0.8} & \textbf{60.7}\std{0.6}\\
        \hline
        \multirow{5}{*}{ActivityNet \cite{krishna2017dense}}
        & NCE & 11.8\std{0.6} & 35.4\std{1.0} & 50.6\std{0.8} & 23.8\std{0.4} & 16.7\std{0.8} & 43.0\std{1.2} & 57.1\std{1.2} & 29.5\std{0.8} & 14.3\std{0.6} & 39.2\std{0.8} & 53.8\std{1.0} & 26.7\std{0.5}\\
        & SHN & 9.9\std{0.9} & 31.2\std{1.3} & 45.2\std{0.9} & 20.9\std{0.9} & 13.7\std{1.1} & 38.5\std{0.9} & 53.4\std{0.9} & 25.9\std{1.0} & 11.8\std{0.9} & 34.9\std{0.8} & 49.3\std{0.7} & 23.4\std{0.9}\\
        & MMS & 12.0\std{0.7} & 35.5\std{1.0} & 49.2\std{0.8} & 23.9\std{0.6} & 16.2\std{0.4} & 42.4\std{0.9} & 56.5\std{1.6} & 28.8\std{0.6} & 14.1\std{0.4} & 39.0\std{0.2} & 52.8\std{1.2} & 26.4\std{0.2}\\
        & AMM & \textbf{17.2}\std{1.1} & \textbf{46.1}\std{1.4} & \textbf{60.0}\std{0.8} & \textbf{30.6}\std{0.6} & \textbf{20.9}\std{1.1} & \textbf{50.1}\std{1.3} & \textbf{62.4}\std{0.8} & \textbf{34.3}\std{0.6} & \textbf{19.1}\std{1.0} & \textbf{48.1}\std{1.2} & \textbf{61.2}\std{0.6} & \textbf{32.5}\std{0.6}\\
        \hline
        \multirow{5}{*}{MSR-VTT \cite{xu2016msr-vtt}}
        & NCE & 20.7\std{0.9} & 51.0\std{0.7} & 66.6\std{1.2} & 35.0\std{0.4} & 30.7\std{1.4} & 65.1\std{0.7} & 78.2\std{1.3} & 46.1\std{1.2} & 25.7\std{1.0} & 58.1\std{0.6} & 72.4\std{1.2} & 40.6\std{0.7}\\
        & SHN & 11.3\std{0.2} & 32.0\std{1.0} & 44.9\std{1.4} & 21.9\std{0.3} & 22.1\std{0.9} & 54.5\std{1.6} & 68.9\std{1.4} & 37.0\std{1.1} & 16.7\std{0.5} & 43.3\std{0.5} & 56.9\std{0.9} & 29.5\std{0.5}\\
        & MMS & 17.6\std{1.1} & 46.5\std{0.9} & 61.6\std{0.9} & 31.5\std{0.6} & 28.3\std{1.1} & 63.1\std{1.4} & 76.1\std{0.9} & 43.8\std{1.1} & 23.0\std{0.9} & 54.8\std{0.6} & 68.9\std{0.7} & 37.6\std{0.6}\\
        & AMM & \textbf{25.7}\std{0.8} & \textbf{61.0}\std{0.8} & \textbf{75.6}\std{0.7} & \textbf{41.6}\std{0.6} & \textbf{32.5}\std{1.5} & \textbf{67.5}\std{1.7} & \textbf{80.1}\std{1.4} & \textbf{48.0}\std{1.2} & \textbf{29.1}\std{0.8} & \textbf{64.2}\std{1.0} & \textbf{77.9}\std{1.0} & \textbf{44.8}\std{0.8}\\
        \hline
        \multirow{5}{*}{S-MiT}
        & NCE & \textbf{33.1}\std{0.9} & \textbf{66.9}\std{1.9} & \textbf{77.6}\std{1.2} & \textbf{47.9}\std{0.7} & 43.0\std{0.8} & \textbf{71.8}\std{0.9} & 80.7\std{1.2} & 55.8\std{0.7} & 38.0\std{0.5} & \textbf{69.3}\std{1.4} & \textbf{79.1}\std{1.1} & 51.8\std{0.6}\\
        & SHN & 23.1\std{1.3} & 55.4\std{1.6} & 69.3\std{1.3} & 37.7\std{1.1} & 41.4\std{1.1} & 70.8\std{0.9} & 79.5\std{1.0} & 54.5\std{0.7} & 32.3\std{0.9} & 63.1\std{1.1} & 74.4\std{1.1} & 46.1\std{0.8}\\
        & MMS & 26.5\std{1.3} & 58.3\std{1.4} & 72.0\std{0.9} & 41.1\std{1.1} & 43.3\std{1.3} & 71.2\std{1.4} & 79.9\std{0.8} & 55.8\std{1.2} & 34.9\std{1.2} & 64.8\std{1.2} & 76.0\std{0.8} & 48.5\std{1.1}\\
        & AMM & \textbf{33.1}\std{0.9} & 65.5\std{1.5} & 76.6\std{1.3} & 47.8\std{1.1} & \textbf{43.8}\std{0.7} & 71.5\std{1.2} & \textbf{80.9}\std{1.6} & \textbf{56.4}\std{0.7} & \textbf{38.4}\std{0.4} & 68.5\std{1.3} & 78.7\std{1.4} & \textbf{52.1}\std{0.8}\\
        \Xhline{0.8pt}
    \end{tabular}}
\caption{\textbf{Loss Function Comparison for Video/Caption Retrieval:}  Models trained on four datasets with different loss functions are compared. 
The proposed AMM loss function consistently achieves the best performance.}
\label{table:loss_comparison}
\end{minipage}
\end{table*}

\begin{table*}[tb]
\begin{minipage}{\linewidth}
    \centering
    \scalebox{0.6}{
    \begin{tabular}{ c | c | c  c  c  c | c  c  c  c | c  c  c  c }
        \Xhline{0.8pt}
        \textbf{Spoken} & \multirow{2}{*}{\textbf{Loss}} & \multicolumn{4}{c|}{\textbf{Caption to Video}} & \multicolumn{4}{c|}{\textbf{Video to Caption}} & \multicolumn{4}{c}{\textbf{Mean}} \\
        \textbf{Caption Model} & & R@1 & R@5 & R@10 & mAP & R@1 & R@5 & R@10 & mAP & R@1 & R@5 & R@10 & mAP \\
        \Xhline{0.8pt}

        \multirow{4}{*}{ResDavenet~\cite{david2020ijcv}}  
        & NCE & 30.7\std{0.6} & 57.1\std{0.6} & 67.6\std{1.0} & 42.9\std{0.8} & 29.3\std{1.0} & 55.8\std{1.2} & 66.2\std{1.4} & 41.8\std{0.9} & 30.0\std{0.8} & 56.4\std{0.9} & 66.9\std{1.2} & 42.3\std{0.8}\\
        & SHN & 30.2\std{1.1} & 56.9\std{0.8} & 66.8\std{0.5} & 42.6\std{1.0} & 31.0\std{1.2} & 57.2\std{0.8} & 67.1\std{0.9} & 43.2\std{1.0} & 30.6\std{1.1} & 57.0\std{0.8} & 67.0\std{0.7} & 42.9\std{1.0}\\
        & MMS & 32.1\std{1.1} & 58.9\std{1.0} & 68.6\std{1.5} & 44.4\std{0.8} & 32.3\std{1.3} & 57.9\std{1.1} & 68.1\std{1.5} & 44.3\std{1.2} & 32.2\std{1.2} & 58.4\std{1.0} & 68.4\std{1.5} & 44.3\std{1.0}\\
        & AMM & \textbf{34.8}\std{1.1} & \textbf{62.0}\std{1.1} & \textbf{70.4}\std{1.2} & \textbf{47.0}\std{1.1} & \textbf{34.6}\std{1.5} & \textbf{60.8}\std{1.6} & \textbf{70.0}\std{0.9} & \textbf{46.8}\std{1.2} & \textbf{34.7}\std{1.2} & \textbf{61.4}\std{1.4} & \textbf{70.2}\std{1.1} & \textbf{46.9}\std{1.1}\\
        \hline
        \multirow{4}{*}{Wav2Vec~\cite{wav2vec}}
        & NCE & 32.6\std{0.7} & 60.4\std{0.8} & 70.3\std{1.6} & 45.3\std{0.8} & 30.9\std{1.0} & 59.6\std{0.9} & 69.8\std{1.1} & 43.9\std{0.8} & 31.8\std{0.7} & 60.0\std{0.8} & 70.0\std{1.3} & 44.6\std{0.8}\\ 
        & SHN & 27.8\std{1.0} & 54.2\std{1.7} & 64.9\std{1.8} & 40.1\std{1.0} & 28.4\std{0.7} & 53.7\std{1.6} & 64.2\std{1.7} & 40.4\std{0.8} & 28.1\std{0.8} & 53.9\std{1.6} & 64.6\std{1.7} & 40.2\std{0.9}\\
        & MMS & 33.6\std{0.6} & 60.5\std{1.2} & \textbf{71.4}\std{1.1} & 46.1\std{0.7} & 33.4\std{1.0} & 60.5\std{1.7} & \textbf{70.3}\std{1.1} & 45.7\std{0.8} & 33.5\std{0.6} & 60.5\std{1.4} & \textbf{70.8}\std{1.1} & 45.9\std{0.7}\\
        & AMM & \textbf{35.0}\std{0.4} & \textbf{61.7}\std{0.9} & 71.0\std{0.9} & \textbf{47.1}\std{0.6} & \textbf{34.7}\std{1.5} & \textbf{61.1}\std{0.9} & 70.2\std{0.9} & \textbf{46.8}\std{1.2} & \textbf{34.8}\std{0.9} & \textbf{61.4}\std{0.9} & 70.6\std{0.8} & \textbf{47.0}\std{0.9}\\
        \hline
        \multirow{4}{*}{ResNet-34} 
        & NCE & 32.2\std{1.3} & 59.7\std{1.4} & 70.3\std{1.3} & 44.8\std{1.1} & 32.8\std{1.8} & 58.8\std{1.3} & 69.2\std{1.9} & 45.1\std{1.4} & 32.5\std{1.4} & 59.2\std{1.3} & 69.7\std{1.5} & 45.0\std{1.2}\\
        & SHN & 32.7\std{1.1} & 60.3\std{1.3} & 71.0\std{1.1} & 45.5\std{1.0} & 33.1\std{1.0} & 60.1\std{1.5} & 70.1\std{1.3} & 45.6\std{0.9} & 32.9\std{1.0} & 60.2\std{1.4} & 70.6\std{1.2} & 45.6\std{0.9}\\
        & MMS & 35.3\std{1.0} & 62.5\std{1.2} & 72.8\std{1.8} & 47.7\std{0.6} & 36.7\std{0.9} & 62.2\std{0.8} & 72.1\std{1.6} & 48.6\std{0.9} & 36.0\std{0.7} & 62.3\std{1.0} & 72.5\std{1.6} & 48.2\std{0.7}\\
        & AMM & \textbf{36.3}\std{0.5} & \textbf{63.9}\std{1.7} & \textbf{73.7}\std{1.6} & \textbf{48.9}\std{0.8} & \textbf{37.5}\std{1.7} & \textbf{63.5}\std{1.9} & \textbf{73.7}\std{1.6} & \textbf{49.6}\std{1.5} & \textbf{36.9}\std{1.1} & \textbf{63.7}\std{1.7} & \textbf{73.7}\std{1.5} & \textbf{49.2}\std{1.2}\\
        \hline
        \multirow{4}{*}{ResNet-50} 
        & NCE & 32.7\std{0.6} & 60.8\std{1.9} & 70.6\std{1.6} & 45.6\std{0.8} & 33.1\std{1.0} & 59.4\std{1.5} & 69.6\std{1.4} & 45.5\std{0.9} & 32.9\std{0.5} & 60.1\std{1.7} & 70.1\std{1.4} & 45.5\std{0.8}\\
        & SHN & 33.9\std{0.6} & 60.1\std{1.4} & 70.9\std{1.3} & 45.8\std{0.7} & 34.0\std{1.2} & 60.6\std{1.8} & 70.1\std{1.4} & 46.0\std{1.1} & 34.0\std{0.8} & 60.3\std{1.5} & 70.5\std{1.3} & 45.9\std{0.8}\\
        & MMS & 37.2\std{0.9} & 65.4\std{0.6} & 75.1\std{1.3} & 50.0\std{0.7} & 37.8\std{1.3} & 64.6\std{1.1} & 74.2\std{0.9} & 50.1\std{1.1} & 37.5\std{1.0} & 65.0\std{0.8} & 74.7\std{1.1} & 50.0\std{0.9}\\
        & AMM & \textbf{39.5}\std{1.3} & \textbf{65.7}\std{1.5} & \textbf{75.5}\std{1.3} & \textbf{51.6}\std{1.1} & \textbf{40.1}\std{0.7} & \textbf{66.3}\std{1.1} & \textbf{74.5}\std{1.2} & \textbf{52.0}\std{0.7} & \textbf{39.8}\std{0.9} & \textbf{66.0}\std{1.2} & \textbf{75.}0\std{1.1} & \textbf{51.8}\std{0.8}\\
        \Xhline{0.8pt}
    \end{tabular}
    }
\caption{\textbf{Spoken Caption Model Comparison:} Models trained with different spoken caption architectures and different loss functions are compared for video/caption retrieval on the S-MiT test set. The proposed AMM loss function consistently achieves the highest performance while ResNet-50 is found to be significantly stronger than the other architectures.}
\label{table:audio_model_comparison_retrieval}
\vspace{5mm}
\scalebox{0.6}{
    \begin{tabular}{ c | c c c c | c c c c | c c c c | c c c c | c c c c }
        \Xhline{0.8pt}
        \multirow{3}{*}{\textbf{Trained On}} & \multicolumn{20}{c}{\textbf{Evaluated On}} \\
        \cline{2-21}
        & \multicolumn{4}{c|}{\textbf{Vatex}} 
        & \multicolumn{4}{c|}{\textbf{ActivityNet}} 
        & \multicolumn{4}{c|}{\textbf{MSR-VTT}} 
        & \multicolumn{4}{c|}{\textbf{S-MiT}} 
        & \multicolumn{4}{c}{\textbf{Mean}}\\  
        & R@1 & R@5 & R@10 & mAP
        & R@1 & R@5 & R@10 & mAP
        & R@1 & R@5 & R@10 & mAP
        & R@1 & R@5 & R@10 & mAP
        & R@1 & R@5 & R@10 & mAP \\
        \Xhline{0.8pt}
        \textbf{Vatex} & \textbf{45.9} & \textbf{79.7} & \textbf{87.5} & \textbf{60.7} & 15.6 & 39.4 & 51.7 & 27.1 & 22.6 & 49.8 & 63.2 & 35.6 & 13.1 & 33.0 & 45.8 & 23.5 & 24.3 & 50.5 & 62.1 & 36.7\\
        \textbf{ActivityNet} & 25.0 & 56.0 & 68.4 & 39.1 & \textbf{19.1} & \textbf{48.1} & \textbf{61.2} & \textbf{32.5} & 15.1 & 37.1 & 50.4 & 26.4 & 9.8 & 28.7 & 40.6 & 19.7 & 17.3 & 42.5 & 55.2 & 29.4\\
        \textbf{MSR-VTT} & 21.0 & 51.3 & 64.8 & 35.1 & 9.9 & 28.3 & 39.7 & 19.6 & 29.1 & 64.2 & \textbf{77.9} & 44.8 & 14.6 & 39.3 & 53.4 & 26.9 & 18.7 & 45.8 & 59.0 & 31.6\\
        \textbf{S-MiT} & 42.7 & 75.4 & 84.2 & 57.1 & 17.6 & 41.6 & 53.8 & 29.2 & \textbf{33.1} & \textbf{64.8} & 77.4 & \textbf{47.6} & \textbf{38.4} & \textbf{68.5} & \textbf{78.7} & \textbf{52.1} & \textbf{33.0} & \textbf{62.6} & \textbf{73.5} & \textbf{46.5}\\
        \Xhline{0.8pt}
    \end{tabular}}
\caption{\textbf{Cross Dataset Evaluation on Video/Caption Retrieval:} Here we compare the generalization performance of models trained on four different datasets for video/caption retrieval.  Each model is trained on a single dataset and we average the evaluation on five 1k video-caption samples from the test set of each other dataset.  We additionally show the mean performance accross datasets.  The S-MiT model shows it generalizes very strongly to the other datasets even beating the MSR-VTT model on its own test set.}
\label{table:cross_evaluation}
\end{minipage}
\end{table*}

\subsection{Adaptive Mean Margin}
\label{sec:approach}
We train our model using the contrastive loss 
with a similar setting to MMS (Equation \ref{eq:margin_nce}). The only difference is that we replace the margin, $M$, with an adaptive margin based on the difference between the similarity of the positive pair and the set of negative pairs in each batch. 


The challenge in using the MMS margin for mini-batch sampled contrastive learning is that the initial margin and growth schedule are difficult to tune for a specific dataset and similarity metric.  Additionally, depending on the sampled pairs in a mini-batch, the margin calculated may be too weak if the positive pair is much more similar than the sampled negative pairs and too strong if it is very similar to the negative pairs.  The approach to monotonically increase the margin during training is meant to address this as the positive and negative pairs will share similar alignment early in training and begin to diverge closer to convergence.  However, variable rates of convergence of different models on different datasets make this growth rate difficult to tune and this approach does not account for differences in the negative samples that appear in different mini-batches.  To address this, we propose an adaptive margin based on relative batch-wise similarity scores.

Class labels have been proposed to be used for generating adaptive margins based on class similarity between positive and negative pairs \cite{adaptivemargin2020cvpr, Liu_2019_CVPR}. Likewise, prior work explored a non-class dependant approach for an adaptive similarity-based margin for human pose estimation \cite{Li_2015_ICCV} where the mean joint error between a positive pose and a hard sampled negative pose was used as a margin with the triplet loss. This adaptively increases the margin when the sampled negative pair is dissimilar to the positive pair in order to maximize the learning signal on less aligned negative samples.  We follow a similar intuition and simply replace $M$ in Equation \ref{eq:margin_nce} with 
\begin{align}
\text{\footnotesize $M_{xy}=\alpha\big(\mathcal{S}(x_i,y_i) - \frac{1}{B-1}\sum^{B}_{j=1}\mathit{I}_{i \neq j}\mathcal{S}(x_i,y_j)\big)$},
\label{eq:adaptive_mean_margin}
\end{align}
where $\alpha$ is a dampening parameter to weight the strength of the margin. 
When $M_{xy}$ in Equation~\ref{eq:adaptive_mean_margin} is applied to Equation~\ref{eq:margin_nce} with  $\alpha=1$, the margin removes the positive pair similarity from the optimization. 
Ablation studies on different alpha values can be found in the supplementary material. 
In practice we use $\alpha=0.5$ in our experiments.

This has the effect of increasing the margin as the difference between the true pair similarity and the similarity of the negative pairs increases. As the training progresses, and the learning approaches convergence, the margin generally increases with the increased separation between positive and negative pair-wise similarities. This also removes the need to tune the margin and growth rate which may have different optimal values for different similarity metrics, batch sizes and datasets. 

We refer to this as an \textbf{Adaptive Mean Margin} (AMM) for contrastive learning and show 
in Section \ref{sec:results} the effect of applying this adaptive margin.

\section{Results}
\label{sec:results}

\subsection{Video/Caption Retrieval}
In Tables~\ref{table:language_model_comparison_retrieval},~\ref{table:loss_comparison} and~\ref{table:audio_model_comparison_retrieval} we show results of R@k recall scores (for $k=1,5,10$) and mean average precision (mAP) on both caption to video and video to caption retrieval. 
Results are averaged over five random sets of 1k video-caption pairs from the test set. Each model in Tables~\ref{table:language_model_comparison_retrieval} and ~\ref{table:loss_comparison} uses the output of a pretrained ASR model, the Google Cloud ASR engine, as input into a trained language model to generate a feature representation for each caption. 
Alternatively, the spoken caption models align visual representations directly from the audio signal without pretrained modules.

Table~\ref{table:language_model_comparison_retrieval} shows the result of using different language models to generate our caption representations from ASR text transcriptions. Each of these models was trained using the proposed AMM loss function described in Section~\ref{sec:approach}. We evaluate the AMM loss in Table~\ref{table:loss_comparison} where we compare the results on the NCE, SHN, MMS and AMM loss functions described in Sections ~\ref{sec:related_loss} and ~\ref{sec:approach} on four different datasets (the proposed Spoken Moments in Time dataset (S-MiT) as well as Vatex-en \cite{Wang_2019_ICCV}, MSR-VTT \cite{xu2016msr-vtt} and ActivityNet Captions\footnote{We used the groundtruth timestamps to get corresponding video clips.} \cite{krishna2017dense}).  
The proposed AMM loss function consistently achieves the best results across each dataset in Table \ref{table:loss_comparison} and the BART language model provides the strongest representations for the retrieval task in Table \ref{table:language_model_comparison_retrieval}.  


Table \ref{table:loss_comparison} shows a comparison of our 
AMM approach to other methods for cross-modal contrastive learning.  We use the BART language model \cite{lewis2019bart} to generate representations of words transcribed from the audio captions via a pretrained ASR model.  Replacing the monotonically increasing margin used in MMS \cite{ilharco2019largescale} with an adaptive margin that scales with the samples in a batch achieves the strongest results.  We observed that as training continues and the margin in MMS continues to grow the training performance begins to degrade.  This is likely due to the margin becoming too large for stable training as described in prior work \cite{Wang_2018_CVPR}.

In Table~\ref{table:audio_model_comparison_retrieval}, we show a comparison of different spoken caption models with different loss functions. The proposed AMM approach beats the other loss functions consistently. 

\subsection{Cross Dataset Evaluation}

To further examine the strength of our proposed Spoken Moments in Time (S-MiT) dataset, we compare the generalization performance of models trained on four different datasets (S-MiT as well as Vatex-en \cite{Wang_2019_ICCV}, MSR-VTT \cite{xu2016msr-vtt} and ActivityNet Captions \cite{krishna2017dense}) for video/caption retrieval (see Table \ref{table:caption_stats} (right) for comparisons of these datasets).  We train each model on a single dataset using the approach described in Section \ref{sec:approach} and evaluate on the test set from each other dataset.  For example, a model trained on Vatex is evaluated on, in addition to its own, the test sets of ActivityNet Captions, MSR-VTT and S-MiT.  We sample five sets of 1k video-caption pairs from each test set.  This allows us to fairly compare results across test sets of different sizes (see supplementary material for full test set results).  Each model in this evaluation was trained using the BART \cite{lewis2019bart} language model and the proposed AMM loss function which was found to give the best results (see Tables \ref{table:language_model_comparison_retrieval}, \ref{table:loss_comparison}).  We evaluate the models using the mean between the video-to-caption and caption-to-video retrieval tasks. We are not able to compare the spoken caption models from Table \ref{table:audio_model_comparison_retrieval} here as the other datasets only include text captions.

In Table \ref{table:cross_evaluation}, we can see that the S-MiT model generalizes better than the other models in spite of the additional noise introduced by the ASR model.  
Additionally, the restriction to 3-second videos in S-MiT does not hinder it's ability to generalize to the much longer videos of the other datasets.

\section{Conclusions}
    In this paper, we have introduced the Spoken Moments in Time dataset which includes 500k pairs of video clips and corresponding spoken descriptions. This new dataset represents the largest video caption dataset available and will serve as a new benchmark for the community. We compared various benchmark models for learning joint representations between captions and videos, and evaluated our approaches on multiple datasets to highlight the strength of the models as well as the ability of models trained on our proposed dataset to generalize to tasks in other datasets.  With these results we are confident that the presented Spoken Moments dataset will have a positive impact on the fields of video understanding and cross-modal learning.

\section{Acknowledgment}
    This work was supported by the MIT-IBM Watson AI Lab as well as the Intelligence Advanced Research Projects Activity (IARPA) via Department of Interior/ Interior Business Center (DOI/IBC) contract number D17PC00341. 

\section{Disclaimer} 
The U.S. Government is authorized to reproduce and distribute reprints for Governmental purposes notwithstanding any copyright annotation thereon. The views and conclusions contained herein are those of the authors and should not be interpreted as necessarily representing the official policies or endorsements, either expressed or implied, of IARPA, DOI/IBC, or the U.S. Government.

{\small
\balance
\bibliographystyle{ieee_fullname}
\bibliography{egbib}
}
\renewcommand\thesection{\Alph{section}}
\setcounter{section}{0}

\clearpage
\section{Annotation}

\begin{figure*}
    \centering
    \includegraphics[width=0.95\linewidth]{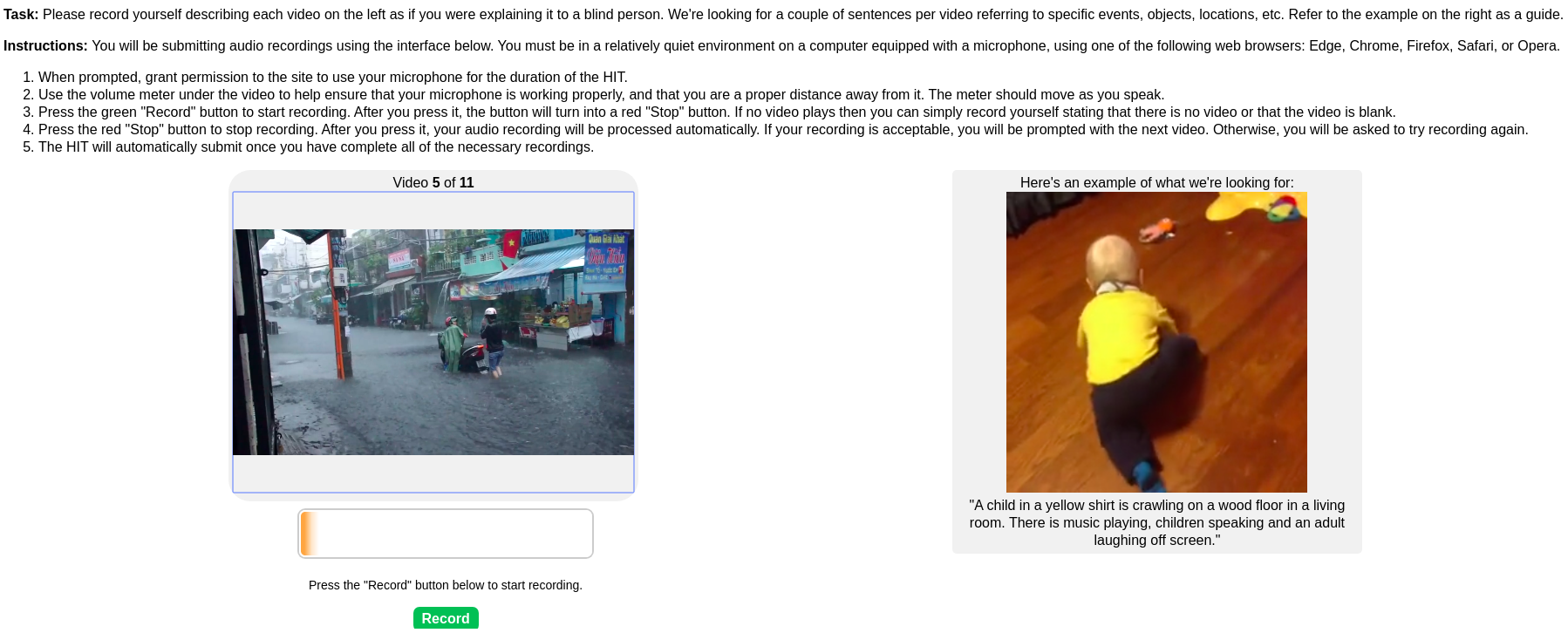}
    \caption{\textbf{Spoken Caption Collection:} Target videos for which descriptions are collected on the left and a video with an example text description is always visible on the right.}
    \label{fig:amt_interface}
\end{figure*}

We follow the approach used to collect the Places Audio Caption dataset \cite{david2016nips,david2020ijcv} and collect audio descriptions of each video in the dataset using Amazon Mechanical Turk (AMT).  In order to ensure that we have a large and diverse dataset, we collect an audio description using AMT for each video in a set of 500k randomly selected videos from the training set and at least two unique descriptions for each video in the 10k videos used for both the validation and test sets.  Each AMT worker is presented with a task of recording themselves describing 10 different videos.  Each video is shown on the left of the screen while a video with an example text description is shown on the right.  This example helps to show the workers the types of descriptions we are looking for and the amount of detail we expect from them.  This example stays on the right side of the screen throughout the task while the target videos on the left cycle as the worker completes each description. Figure \ref{fig:amt_interface} shows an example of this interface with an example video and caption on the right and a target video on the left.  Below each target description is a button that allows the worker to start recording their voice as they describe the video.  Once they press this button, the video is removed from the screen and the recording is started.  We block the worker from seeing the video while recording the description to ensure that the recordings are concise and pertain only to the important events highlighted in their memory.  We use the Google Cloud ASR engine to verify the quality of each recorded description and flag AMT workers for poor performance.  This is done by checking that the generated text has at least five words, is unique (some bots repeat pre-recorded audio to trick the system) and that the audio is at least three seconds long.  If any of these checks fail we don't let the worker continue to the next video until they record a new description that passes our checks.  Once the descriptions are recorded, we periodically sample videos to check the quality of the audio paired with the ASR to ensure they match the videos and have an appropriate level of detail. If these checks fail, we flag the workers that recorded the descriptions, don't allow them to record in the future and recheck all of their recorded data.  This process allows us to ensure a strong level of quality in our collected spoken captions. 
Examples of some of the videos and corresponding text transcriptions of the descriptions we collected can be seen in Figure \textcolor{red}{1}.
\section{Implementation Details}
We train each model on a server with 8 24GB Titan RTX cards using a mini-batch size of 2048 for 100 epochs. We examine the effect of the mini-batch size on learning in the next section. We take the best parameters as evaluated on the evaluation set of the training dataset after each epoch. We repeat this process for two phases of training.  First we freeze the visual backbone models and train only the projection heads (including the full caption model for the spoken models) and then, in a second round, allow the full visual model to train as well.  We keep the language and ASR components frozen for the language caption models and reserve fine-tuning these components for future work.  
For model training, we use an Adam \cite{Kingma2015AdamAM} optimizer where a fixed learning rate of $0.001$ and $0.00001$ are set for the first and the second round model training, respectively.


\section{Ablation Studies}
\label{sec:ablation_studies}

\begin{table*}[tb]
    \centering
    \scalebox{0.67}{
    \begin{tabular}{ c | c | c  c  c  c | c  c  c  c | c  c  c  c }
        \hline
        \multirow{2}{*}{\textbf{Dataset}} & \textbf{Pretrained TSM} & \multicolumn{4}{c|}{\textbf{Caption to Video}} & \multicolumn{4}{c|}{\textbf{Video to Caption}} & \multicolumn{4}{c}{\textbf{Mean}} \\
        & \textbf{Dataset} & R@1 & R@5 & R@10 & mAP & R@1 & R@5 & R@10 & mAP & R@1 & R@5 & R@10 & mAP \\
\hline
\textbf{Vatex} \cite{Wang_2019_ICCV}
        & \textbf{Kinetics} & 39.6\std{1.0} & 77.5\std{1.5} & 87.2\std{1.0} & 55.9\std{0.8} & 46.4\std{0.6} & \textbf{82.1}\std{1.0} & \textbf{90.2}\std{1.2} & \textbf{61.9}\std{0.6} & 43.0\std{0.7} & 79.8\std{1.1} & \textbf{88.7}\std{1.0} & 58.9\std{0.7}\\
        & \textbf{S-MiT} & \textbf{47.4}\std{1.1} & \textbf{81.5}\std{0.7} & \textbf{89.0}\std{1.1} & \textbf{62.3}\std{0.6} & 43.1\std{0.9} & 78.3\std{0.6} & 86.2\std{0.3} & 58.5\std{0.5} & \textbf{45.3}\std{0.8} & \textbf{79.9}\std{0.4} & 87.6\std{0.6} & \textbf{60.4}\std{0.5}\\
        \hline
\textbf{ActivityNet} \cite{krishna2017dense}
        & \textbf{Kinetics} & \textbf{18.7}\std{1.0} & \textbf{45.6}\std{0.9} & 57.2\std{1.4} & \textbf{31.0}\std{0.7} & \textbf{20.8}\std{0.8} & \textbf{50.1}\std{1.4} & \textbf{61.8}\std{1.3} & \textbf{34.1}\std{0.4} & \textbf{19.8}\std{0.8} & \textbf{47.8}\std{0.9} & \textbf{59.5}\std{1.0} & \textbf{32.5}\std{0.4}\\
        & \textbf{M-MiT} & 16.1\std{1.7} & 44.0\std{1.0} & \textbf{57.5}\std{1.7} & 29.3\std{1.0} & 19.0\std{1.3} & 48.2\std{0.9} & 61.0\std{1.1} & 32.5\std{0.8} & 17.6\std{1.5} & 46.1\std{0.7} & 59.2\std{1.4} & 30.9\std{0.8}\\
\hline
\textbf{MSR-VTT} \cite{xu2016msr-vtt}
        & \textbf{Kinetics} & 17.6\std{1.3} & 48.9\std{1.8} & 65.6\std{1.2} & \textbf{31.6}\std{1.3} & 25.5\std{0.7} & 59.7\std{1.8} & 74.1\std{1.6} & 40.6\std{0.9} & 21.6\std{0.8} & 54.3\std{1.4} & 69.8\std{1.4} & 36.1\std{0.9}\\
        & \textbf{M-MiT} & \textbf{20.7}\std{0.5} & \textbf{54.2}\std{0.9} & \textbf{70.6}\std{1.0} & 30.5\std{0.4} & \textbf{31.3}\std{1.1} & \textbf{61.0}\std{1.0} & \textbf{75.0}\std{0.9} & \textbf{40.9}\std{0.8} & \textbf{24.0}\std{0.6} & \textbf{57.6}\std{0.6} & \textbf{72.8}\std{0.8} & \textbf{37.7}\std{0.4}\\
\hline
\textbf{S-MiT}
        & \textbf{Kinetics} & 27.6\std{1.4} & 57.5\std{2.4} & 70.4\std{1.9} & 41.3\std{1.7} & 37.2\std{2.3} & 65.0\std{1.7} & 75.2\std{1.5} & 50.0\std{1.7} & 32.4\std{1.8} & 61.3\std{2.0} & 72.8\std{1.6} & 45.7\std{1.7}\\
        & \textbf{M-MiT} & \textbf{29.8}\std{2.5} & \textbf{60.6}\std{2.4} & \textbf{72.2}\std{1.9} & \textbf{44.0}\std{2.2} & \textbf{39.4}\std{2.1} & \textbf{68.0}\std{2.0} & \textbf{77.5}\std{1.8} & \textbf{52.3}\std{2.0} & \textbf{34.6}\std{2.1} & \textbf{64.3}\std{2.2} & \textbf{74.9}\std{1.8} & \textbf{48.2}\std{2.0}\\
\hline
    \end{tabular}}
\caption{\textbf{Comparison of different Pretrained TSM models on multiple datasets using AMM and Bart}}
\label{table:pretrained_comparison}
\vspace{5mm}
    \centering
    \scalebox{0.67}{
    \begin{tabular}{ c | c  c  c  c | c  c  c  c | c  c  c  c }
        \hline
        \multirow{2}{*}{\textbf{Visual Base Model}} & \multicolumn{4}{c|}{\textbf{Caption to Video}} & \multicolumn{4}{c|}{\textbf{Video to Caption}} & \multicolumn{4}{c}{\textbf{Mean}} \\
        & R@1 & R@5 & R@10 & mAP & R@1 & R@5 & R@10 & mAP & R@1 & R@5 & R@10 & mAP \\
\hline
\textbf{TSM Kinetics}
        & 20.2\std{1.1} & 47.9\std{2.3} & 61.0\std{0.8} & 33.2\std{1.1} & 28.2\std{1.5} & 54.9\std{1.5} & 67.1\std{1.6} & 40.8\std{1.6} & 24.2\std{1.1} & 51.4\std{1.9} & 64.0\std{1.0} & 37.0\std{1.3}\\
\textbf{TSM M-MiT}
        & 19.7\std{1.1} & 48.6\std{2.0} & 61.9\std{1.6} & 33.5\std{1.3} & 28.4\std{1.4} & 58.0\std{2.5} & 69.2\std{1.9} & 41.9\std{1.4} & 24.1\std{1.2} & 53.3\std{2.1} & 65.6\std{1.7} & 37.7\std{1.4}\\
\textbf{ResNet-152 ImageNet (2D)}
        & 24.2\std{2.4} & 53.6\std{1.8} & 66.5\std{2.1} & 37.9\std{2.0} & 32.9\std{2.1} & 61.7\std{1.6} & 71.6\std{1.0} & 45.9\std{1.8} & 28.5\std{2.2} & 57.7\std{1.7} & 69.1\std{1.5} & 41.9\std{1.9}\\
\textbf{TSM Kinetics + 2D}
         & 27.6\std{1.4} & 57.5\std{2.4} & 70.4\std{1.9} & 41.3\std{1.7} & 37.2\std{2.3} & 65.0\std{1.7} & 75.2\std{1.5} & 50.0\std{1.7} & 32.4\std{1.8} & 61.3\std{2.0} & 72.8\std{1.6} & 45.7\std{1.7}\\
\textbf{TSM M-MiT + 2D}
        & \textbf{29.8}\std{2.5} & \textbf{60.6}\std{2.4} & \textbf{72.2}\std{1.9} & \textbf{44.0}\std{2.2} & \textbf{39.4}\std{2.1} & \textbf{68.0}\std{2.0} & \textbf{77.5}\std{1.8} & \textbf{52.3}\std{2.0} & \textbf{34.6}\std{2.1} & \textbf{64.3}\std{2.2} & \textbf{74.9}\std{1.8} & \textbf{48.2}\std{2.0}\\
\hline
    \end{tabular}}
\caption{\textbf{Comparison of different visual base model combinations on S-MiT using AMM and Bart}}
\label{table:visual_comparison}

\vspace{5mm}
    \centering
    \scalebox{0.67}{
    \begin{tabular}{ c | c  c  c  c | c  c  c  c | c  c  c  c }
        \hline
        \multirow{2}{*}{\textbf{Batch Size}} & \multicolumn{4}{c|}{\textbf{Caption to Video}} & \multicolumn{4}{c|}{\textbf{Video to Caption}} & \multicolumn{4}{c}{\textbf{Mean}} \\
        & R@1 & R@5 & R@10 & mAP & R@1 & R@5 & R@10 & mAP & R@1 & R@5 & R@10 & mAP \\
\hline
\textbf{512}
        & 27.2\std{1.6} & 57.4\std{1.3} & 69.4\std{1.0} & 41.0\std{1.5} & 35.5\std{2.5} & 64.0\std{1.4} & 74.4\std{1.1} & 48.4\std{2.1} & 31.3\std{1.9} & 60.7\std{1.3} & 71.9\std{1.0} & 44.7\std{1.7}\\
\textbf{1024}
        & 27.8\std{2.0} & 57.7\std{1.4} & 69.8\std{1.2} & 41.5\std{1.9} & 36.5\std{2.8} & 65.6\std{1.4} & 75.2\std{1.7} & 49.7\std{2.0} & 32.2\std{2.3} & 61.7\std{1.4} & 72.5\std{1.3} & 45.6\std{1.9}\\
\textbf{2048}
        & \textbf{29.8}\std{2.5} & \textbf{60.6}\std{2.4} & \textbf{72.2}\std{1.9} & \textbf{44.0}\std{2.2} & \textbf{39.4}\std{2.1} & \textbf{68.0}\std{2.0} & \textbf{77.5}\std{1.8} & \textbf{52.3}\std{2.0} & \textbf{34.6}\std{2.1} & \textbf{64.3}\std{2.2} & \textbf{74.9}\std{1.8} & \textbf{48.2}\std{2.0}\\
\textbf{4096}
        & 29.2\std{2.7} & 58.4\std{1.6} & 70.8\std{1.9} & 42.8\std{2.3} & \textbf{39.4}\std{2.3} & 66.6\std{1.8} & 75.7\std{1.4} & 51.8\std{1.9} & 34.3\std{2.3} & 62.5\std{1.6} & 73.3\std{1.6} & 47.3\std{2.0}\\
\hline
    \end{tabular}}
\caption{\textbf{Comparison of different batch sizes on S-MiT using AMM and Bart}}
\label{table:batch_comparison}

\vspace{5mm}
    \centering
    \scalebox{0.67}{
    \begin{tabular}{ c | c  c  c  c | c  c  c  c | c  c  c  c }
        \hline
        \multirow{2}{*}{\textbf{Projection Size}} & \multicolumn{4}{c|}{\textbf{Caption to Video}} & \multicolumn{4}{c|}{\textbf{Video to Caption}} & \multicolumn{4}{c}{\textbf{Mean}} \\
        & R@1 & R@5 & R@10 & mAP & R@1 & R@5 & R@10 & mAP & R@1 & R@5 & R@10 & mAP \\
\hline
\textbf{1024}
        & 27.4\std{1.8} & 56.6\std{1.6} & 69.5\std{0.9} & 41.1\std{1.5} & 38.6\std{1.6} & 66.6\std{1.1} & 76.3\std{1.3} & 51.3\std{1.2} & 33.0\std{1.6} & 61.6\std{1.3} & 72.9\std{1.0} & 46.2\std{1.3}\\
\textbf{2048}
        & 27.8\std{1.8} & 57.4\std{2.0} & 69.2\std{1.5} & 41.5\std{1.8} & 38.4\std{2.1} & 65.9\std{1.4} & 75.6\std{1.5} & 51.1\std{1.6} & 33.1\std{1.9} & 61.6\std{1.6} & 72.4\std{1.4} & 46.3\std{1.7}\\
\textbf{4096}
        & \textbf{29.8}\std{2.5} & \textbf{60.6}\std{2.4} & \textbf{72.2}\std{1.9} & \textbf{44.0}\std{2.2} & \textbf{39.4}\std{2.1} & \textbf{68.0}\std{2.0} & \textbf{77.5}\std{1.8} & \textbf{52.3}\std{2.0} & \textbf{34.6}\std{2.1} & \textbf{64.3}\std{2.2} & \textbf{74.9}\std{1.8} & \textbf{48.2}\std{2.0}\\
\textbf{8192}
        & 29.4\std{2.0} & 58.0\std{2.3} & 70.3\std{1.2} & 42.6\std{1.8} & 38.5\std{2.4} & 66.1\std{2.1} & 76.1\std{1.5} & 51.2\std{2.1} & 33.9\std{2.2} & 62.0\std{2.2} & 73.2\std{1.3} & 46.9\std{1.9}\\
\hline
    \end{tabular}}
\caption{\textbf{Comparison of different projection sizes on S-MiT using AMM and Bart}}
\label{table:projection_comparison}

\vspace{5mm}
    \centering
    \scalebox{0.67}{
    \begin{tabular}{ c | c  c  c  c | c  c  c  c | c  c  c  c }
        \hline
        \multirow{2}{*}{\textbf{Sampling}} & \multicolumn{4}{c|}{\textbf{Caption to Video}} & \multicolumn{4}{c|}{\textbf{Video to Caption}} & \multicolumn{4}{c}{\textbf{Mean}} \\
        & R@1 & R@5 & R@10 & mAP & R@1 & R@5 & R@10 & mAP & R@1 & R@5 & R@10 & mAP \\
\hline
\textbf{N}
        & 28.1\std{1.1} & 57.5\std{2.0} & 69.8\std{1.4} & 41.8\std{1.3} & 39.1\std{1.3} & 66.5\std{2.0} & 76.3\std{1.8} & 51.5\std{1.4} & 33.6\std{1.1} & 62.0\std{1.9} & 73.0\std{1.5} & 46.7\std{1.3}\\
\textbf{Y}
        & \textbf{29.8}\std{2.5} & \textbf{60.6}\std{2.4} & \textbf{72.2}\std{1.9} & \textbf{44.0}\std{2.2} & \textbf{39.4}\std{2.1} & \textbf{68.0}\std{2.0} & \textbf{77.5}\std{1.8} & \textbf{52.3}\std{2.0} & \textbf{34.6}\std{2.1} & \textbf{64.3}\std{2.2} & \textbf{74.9}\std{1.8} & \textbf{48.2}\std{2.0}\\
\hline
    \end{tabular}}
\caption{\textbf{Comparison of sampling approach on S-MiT using AMM and Bart}}
\label{table:sampling_comparison}
\vspace{5mm}

    \centering
    \scalebox{0.67}{
    \begin{tabular}{ c | c  c  c  c | c  c  c  c | c  c  c  c }
        \hline
        \multirow{2}{*}{\textbf{$\alpha$}} & \multicolumn{4}{c|}{\textbf{Caption to Video}} & \multicolumn{4}{c|}{\textbf{Video to Caption}} & \multicolumn{4}{c}{\textbf{Mean}} \\
        & R@1 & R@5 & R@10 & mAP & R@1 & R@5 & R@10 & mAP & R@1 & R@5 & R@10 & mAP \\
\hline
\textbf{0.1}
        & 29.3\std{1.4} & 60.0\std{1.2} & \textbf{72.7}\std{1.4} & 43.4\std{1.2} & 39.2\std{1.8} & 66.2\std{1.5} & 77.0\std{1.3} & 51.7\std{1.6} & 34.2\std{1.5} & 63.1\std{1.3} & 74.8\std{1.1} & 47.5\std{1.4}\\
\textbf{0.2}
        & 28.4\std{1.2} & 58.1\std{1.9} & 70.9\std{1.5} & 42.3\std{1.4} & 39.3\std{1.6} & 67.4\std{1.5} & 77.0\std{1.8} & 52.0\std{1.4} & 33.9\std{1.4} & 62.7\std{1.7} & 74.0\std{1.5} & 47.2\std{1.4}\\
\textbf{0.3}
        & 27.1\std{2.5} & 58.9\std{2.9} & 71.5\std{2.2} & 41.6\std{2.3} & 38.5\std{2.4} & 67.1\std{1.0} & 76.6\std{1.9} & 51.5\std{1.9} & 32.8\std{2.3} & 63.0\std{1.9} & 74.0\std{1.9} & 46.5\std{2.1}\\
\textbf{0.4}
        & 28.1\std{1.1} & 58.1\std{2.1} & 69.8\std{2.3} & 41.9\std{1.5} & 38.8\std{2.4} & 66.9\std{1.2} & 75.8\std{1.5} & 51.5\std{1.8} & 33.5\std{1.8} & 62.5\std{1.6} & 72.8\std{1.7} & 46.7\std{1.6}\\
\textbf{0.5} &
        \textbf{29.8}\std{2.5} & \textbf{60.6}\std{2.4} & 72.2\std{1.9} & \textbf{44.0}\std{2.2} & \textbf{39.4}\std{2.1} & \textbf{68.0}\std{2.0} & \textbf{77.5}\std{1.8} & \textbf{52.3}\std{2.0} & \textbf{34.6}\std{2.1} & \textbf{64.3}\std{2.2} & \textbf{74.9}\std{1.8} & \textbf{48.2}\std{2.0}\\
\textbf{0.6}
        & 28.1\std{2.1} & 59.1\std{2.3} & 71.3\std{2.1} & 42.3\std{1.9} & 38.3\std{1.9} & 67.1\std{1.6} & 76.6\std{1.7} & 51.4\std{1.6} & 33.2\std{1.9} & 63.1\std{1.8} & 73.9\std{1.8} & 46.9\std{1.7}\\
\textbf{0.7}
        & 28.9\std{1.5} & 59.2\std{1.3} & 70.8\std{1.3} & 42.8\std{1.4} & 38.9\std{1.7} & 66.3\std{1.4} & 76.0\std{1.5} & 51.3\std{1.5} & 33.9\std{1.6} & 62.7\std{1.4} & 73.4\std{1.3} & 47.1\std{1.4}\\
\textbf{0.8}
        & 29.0\std{1.9} & 59.2\std{2.4} & 70.7\std{1.4} & 42.8\std{1.9} & 38.3\std{2.1} & 66.3\std{1.6} & 75.9\std{1.5} & 51.1\std{1.7} & 33.6\std{1.9} & 62.8\std{1.9} & 73.3\std{1.3} & 46.9\std{1.7}\\
\textbf{0.9}
        & 27.7\std{2.1} & 57.0\std{2.5} & 68.2\std{2.0} & 41.1\std{2.2} & 37.5\std{2.6} & 64.6\std{2.4} & 74.1\std{1.5} & 49.8\std{2.2} & 32.6\std{2.4} & 60.8\std{2.4} & 71.2\std{1.7} & 45.5\std{2.2}\\
\hline
    \end{tabular}}
\caption{\textbf{Comparison of different dampening multipliers, $\alpha$, in AMM on S-MiT using Bart}}
\label{table:dampening_effect}
\scalebox{0.67}{
    \begin{tabular}{ c | c c c c | c c c c | c c c c | c c c c | c c c c }
        \Xhline{0.8pt}
        \multirow{3}{*}{\textbf{Trained On}} & \multicolumn{20}{c}{\textbf{Evaluated On}} \\
        \cline{2-21}
        & \multicolumn{4}{c|}{\textbf{Vatex}} 
        & \multicolumn{4}{c|}{\textbf{ActivityNet}} 
        & \multicolumn{4}{c|}{\textbf{MSR-VTT}} 
        & \multicolumn{4}{c|}{\textbf{S-MiT}} 
        & \multicolumn{4}{c}{\textbf{Mean}}\\  
        & R@1 & R@5 & R@10 & mAP
        & R@1 & R@5 & R@10 & mAP
        & R@1 & R@5 & R@10 & mAP
        & R@1 & R@5 & R@10 & mAP
        & R@1 & R@5 & R@10 & mAP \\
        \Xhline{0.8pt}
        \textbf{Vatex} &       \textbf{19.8} & \textbf{48.4} & \textbf{63.7} & \textbf{33.4} & 1.5 & 5.2 & 8.6 & 4.2 & 10.3 & 28.7 & 39.3 & 19.8 & 7.1 & 20.2 & 28.6 & 14.4 & 9.7 & 25.6 & 35.1 & 18.0\\
        \textbf{ActivityNet} & 12.1 & 33.3 & 46.8 & 23.0 & 2.0 & 7.3 & 12.0 & 5.6 & 7.5 & 22.1 & 31.2 & 15.4 & 4.9 & 15.6 & 24.1 & 11.4 & 6.6 & 19.6 & 28.5 & 13.9\\
        \textbf{MSR-VTT} &      6.5 & 19.2 & 28.8 & 13.8 & 1.3 & 4.6 & 7.8 & 3.7 & 11.8 & 33.9 & 48.2 & 23.2 & 8.0 & 23.6 & 34.3 & 16.4 & 6.9 & 20.3 & 29.8 & 14.3\\
        \textbf{S-MiT} &       19.4 & 44.6 & 57.7 & 31.7 & \textbf{2.7} & \textbf{8.4} & \textbf{13.6} & \textbf{6.5} & \textbf{17.3} & \textbf{39.8} & \textbf{51.8} & \textbf{28.4} & \textbf{25.8} & \textbf{52.8} & \textbf{64.7} & \textbf{38.5} & \textbf{16.3} & \textbf{36.4} & \textbf{47.0} & \textbf{26.3}\\
        \Xhline{0.8pt}
    \end{tabular}}
\caption{\textbf{Cross Dataset Evaluation on Video/Caption Retrieval on Full Test Set}}
\label{table:cross_evaluation_full}
\end{table*}

In Tables~\ref{table:pretrained_comparison},~\ref{table:visual_comparison},~\ref{table:batch_comparison},~\ref{table:projection_comparison}, and~\ref{table:sampling_comparison}, we show several ablation studies.  Unless otherwise listed in the table we use the proposed AMM loss function with the BART \cite{lewis2019bart} language model as part of the language caption model described in Section \textcolor{red}{4.2.1} for each experiment. Results are averaged over five rounds with a single random batch of 1k caption-video pairs from the test set.  Due to the increased computation demand of these studies we freeze the base models and train the projection heads for alignment.  We use the best model settings found in this analysis to train the full models with results reported in Section \textcolor{red}{5}.

Table~\ref{table:pretrained_comparison} shows the effect of using two different pretrained temporal shift \cite{Lin_2019_ICCV} video models on four different datasets in order to choose the most appropriate base models (Multi-Moments in Time (M-MiT) \cite{monfort2019multimoments} or Kinetics \cite{kay2017kinetics}).  Here we use the BART language model and the proposed AMM loss function as described in Section \textcolor{red}{4} as this combination gave us the best results on each dataset. 

Table \ref{table:visual_comparison} compares the effect of the video model (TSM) trained for action recognition and the 2D model trained for object recognition.  Most captions reference both objects and actions in a video with an average of 4.37 nouns used per caption compared to 1.58 verbs. The strength of the 2D obect model makes sense when we consider this prevalence of nouns in the captions.  The combination of the TSM model trained on M-MiT \cite{monfort2019multimoments} and the 2D models trained on ImageNet \cite{krizhevsky2012imagenet} provided the best performance when used with the model described in Section \textcolor{red}{4}.

In Tables \ref{table:batch_comparison} and \ref{table:projection_comparison} we compare the the effect of the batch size and projection size on the performance of the S-MiT model described in Section~\textcolor{red}{4} in order to validate our choice of a 2048 batch and a 4096 projection.  Similarly, Table \ref{table:sampling_comparison} shows the effect of using the caption sampling approach for the transcription model as described in Section~\textcolor{red}{4.2.1}. In Table~\ref{table:dampening_effect}, we explore different dampening parameters.



\section{Cross Dataset Generalization}

In Table \ref{table:cross_evaluation_full}, we expand on Table \textcolor{red}{4} and compare the generalization performance of models trained on four different datasets (S-MiT as well as Vatex-en \cite{Wang_2019_ICCV}, MSR-VTT \cite{xu2016msr-vtt} and ActivityNet Captions \cite{krishna2017dense}) for video/caption retrieval on their full test sets.  In Table \textcolor{red}{4} we ran the comparison on five samples of 1k video-caption pairs to be consistent on evaluating across different size test sets.  Here we evaluate on the full test set of each dataset to provide a baseline for each test set.  The strength of the model trained on S-MiT is even more evident here as it achieves higher results on the test sets of both ActivityNet and MSR-VTT than the models trained on those datasets.  It even comes very close to the performance of the Vatex model on the Vatex test set.  This shows that the scale and diversity of the S-MiT dataset is highly beneficial to training robust models.
\section{Qualitative Results}
In Tables~\ref{table:analysis_c2i} and~\ref{table:analysis_i2c}, we show the top five retrieval results for some examples from the Spoken Moments dataset. For this analysis, we use the language caption model described in Section~\textcolor{red}{4.2.1} with the BART~\cite{lewis2019bart} language model and the proposed AMM loss function. 
Table~\ref{table:analysis_c2i} shows the top five retrieved captions given a query video, while Table~\ref{table:analysis_i2c} shows the top five retrieved videos given a query caption. Blue boxes indicate the ground-truth results. 

Our model retrieves results by recognizing key objects or environments in the videos. For example, in Table~\ref{table:analysis_c2i} (c), \textit{lettuce} is distinguished from the other vegetables. In Table~\ref{table:analysis_i2c} (f), the model not only \textit{recognizes the planets} in space but also \textit{understands that they are crashing into each other}. 
Some of the examples show that the top retrieval result is not the ground-truth. However, as we can see, the top predictions are typically still a strong match for the queries, as in (e), (i) in Table~\ref{table:analysis_c2i} and (a), (b) in Table~\ref{table:analysis_i2c}.

For this demonstration, we use transcribed words from the audio captions using a pretrained ASR model. Noise in these transcriptions may contribute to some errors. In the future, we plan to investigate jointly training a pre-trained ASR model, and language model, with the video model to improve our performance.

\begin{table*}
\centering
\scalebox{0.85}{
    \begin{tabular}{ c | P{2.5cm}  P{2.5cm} P{2.5cm} P{2.5cm} P{2.5cm} P{2.5cm}}
        \hline 
        
        \multirow{2}{*}{} & \multirow{2}{*}{\textbf{Query}} & \multicolumn{5}{c}{\textbf{Retrieval Results}} \\
        \cline{3-7}
        & & R@1 & R@2 & R@3 & R@4 & R@5 \\ 
        \hline 
        (a) & \multicolumn{6}{c}{\includegraphics[width=1.0\linewidth, trim={0 0 41cm 0}, clip]{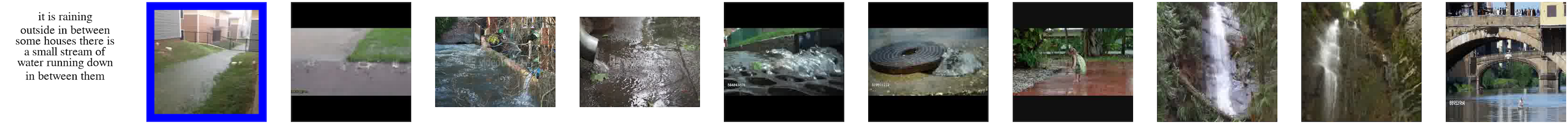}} \\ \hline
        (b) & \multicolumn{6}{c}{\includegraphics[width=1.0\linewidth, trim={0 0 41cm 0}, clip]{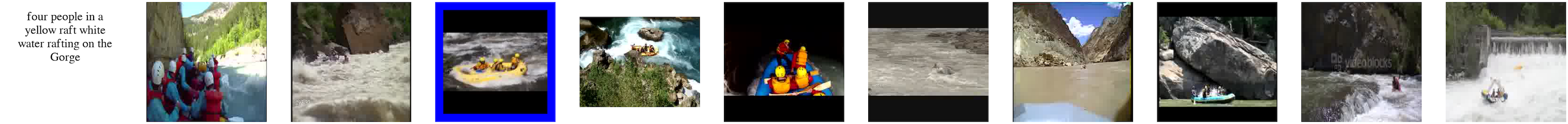}} \\ \hline
        (c) & \multicolumn{6}{c}{\includegraphics[width=1.0\linewidth, trim={0 0 41cm 0}, clip]{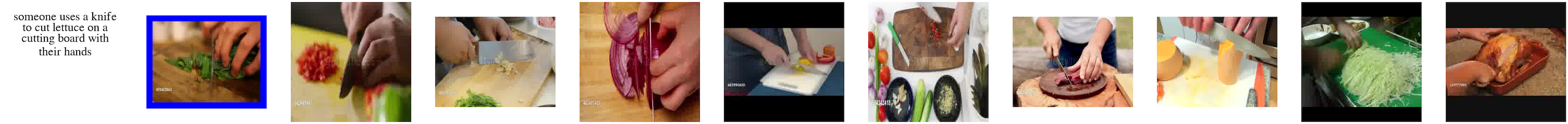}} \\ \hline
        (d) & \multicolumn{6}{c}{\includegraphics[width=1.0\linewidth, trim={0 0 41cm 0}, clip]{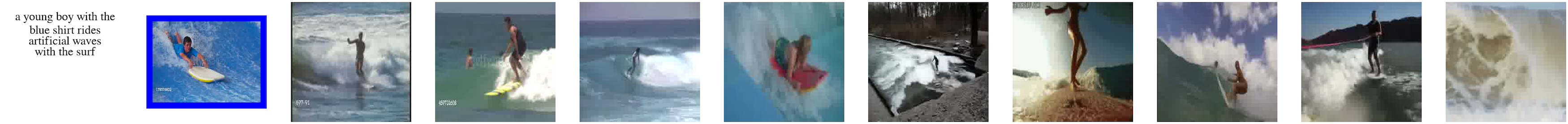}} \\ \hline
        (e) & \multicolumn{6}{c}{\includegraphics[width=1.0\linewidth, trim={0 0 41cm 0}, clip]{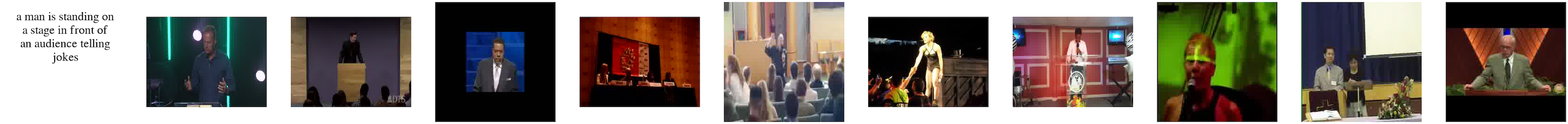}} \\ \hline
        (f) & \multicolumn{6}{c}{\includegraphics[width=1.0\linewidth, trim={0 0 41cm 0}, clip]{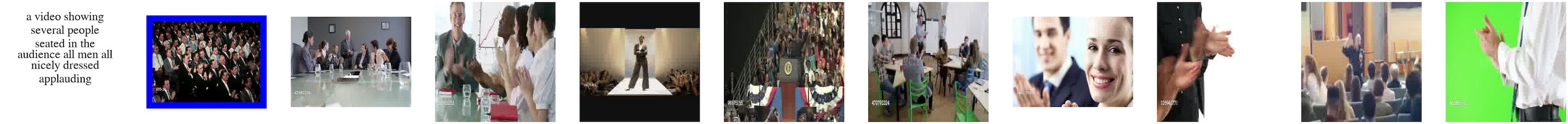}} \\ \hline
        (g) & \multicolumn{6}{c}{\includegraphics[width=1.0\linewidth, trim={0 0 41cm 0}, clip]{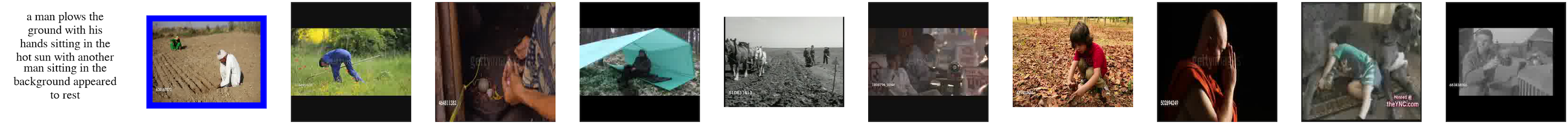}} \\ \hline
        (h) & \multicolumn{6}{c}{\includegraphics[width=1.0\linewidth, trim={0 0 41cm 0}, clip]{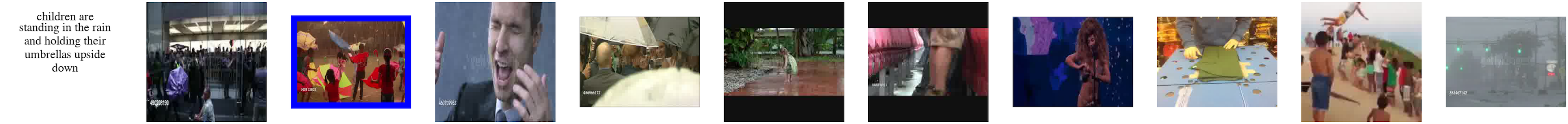}} \\ \hline
        (i) & \multicolumn{6}{c}{\includegraphics[width=1.0\linewidth, trim={0 0 41cm 0}, clip]{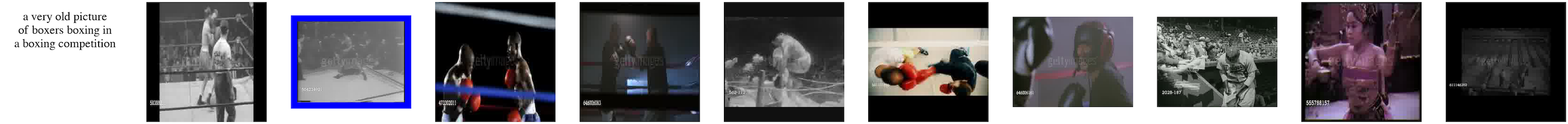}} \\
        \hline 
    \end{tabular}}
\caption{\textbf{Spoken Moments Examples of Caption to Video Retrieval Results:} Given a query caption, we show five top retrieved captions where words transcribed from the audio captions using a pretrained ASR model are used as a caption. We use a BART model trained with the AMM loss function on the S-MiT dataset. Blue indicates the ground-truth results.} 
\label{table:analysis_c2i}
\end{table*}

\begin{table*}
\centering
\scalebox{0.85}{
    \begin{tabular}{ c | P{2.5cm}  P{2.5cm} P{2.5cm} P{2.5cm} P{2.5cm} P{2.5cm}}
        \hline 
        
        \multirow{2}{*}{} & \multirow{2}{*}{\textbf{Query}} & \multicolumn{5}{c}{\textbf{Retrieval Results}} \\
        \cline{3-7}
        & & R@1 & R@2 & R@3 & R@4 & R@5 \\ 
        \hline 
        (a) & \multicolumn{6}{c}{\includegraphics[width=1.0\linewidth, trim={0 0 41cm 0}, clip]{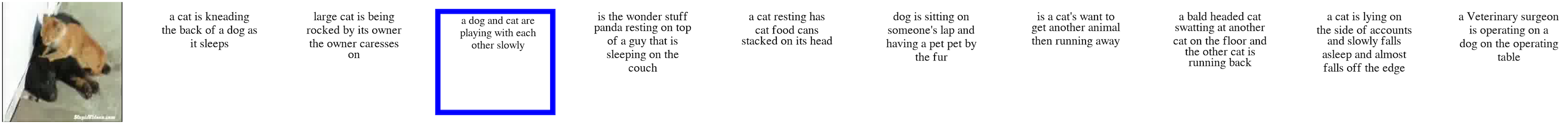}} \\ \hline
        (b) & \multicolumn{6}{c}{\includegraphics[width=1.0\linewidth, trim={0 0 41cm 0}, clip]{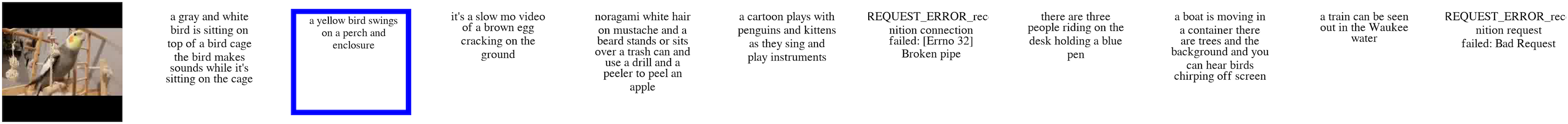}} \\ \hline
        (c) & \multicolumn{6}{c}{\includegraphics[width=1.0\linewidth, trim={0 0 41cm 0}, clip]{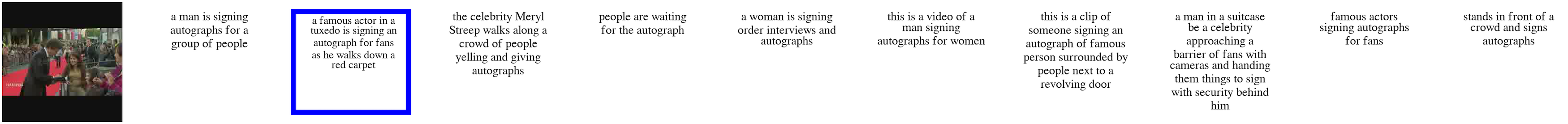}} \\ \hline
        (d) & \multicolumn{6}{c}{\includegraphics[width=1.0\linewidth, trim={0 0 41cm 0}, clip]{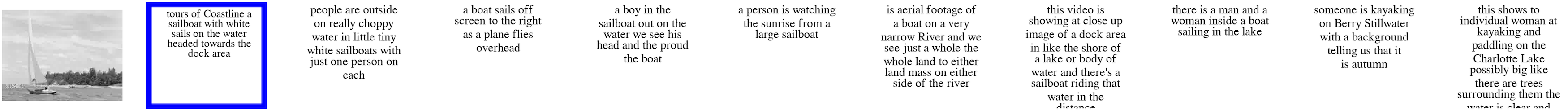}} \\ \hline
        (e) & \multicolumn{6}{c}{\includegraphics[width=1.0\linewidth, trim={0 0 41cm 0}, clip]{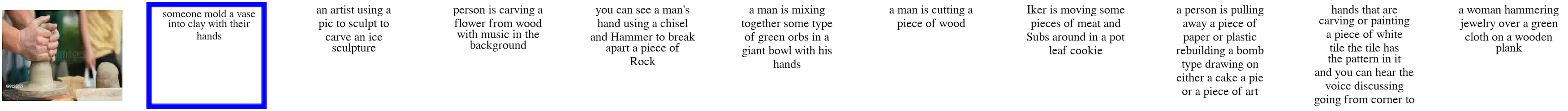}} \\ \hline
        (f) & \multicolumn{6}{c}{\includegraphics[width=1.0\linewidth, trim={0 0 41cm 0}, clip]{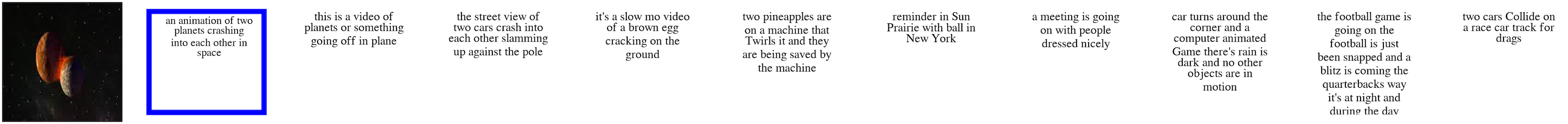}} \\ \hline
        (g) & \multicolumn{6}{c}{\includegraphics[width=1.0\linewidth, trim={0 0 41cm 0}, clip]{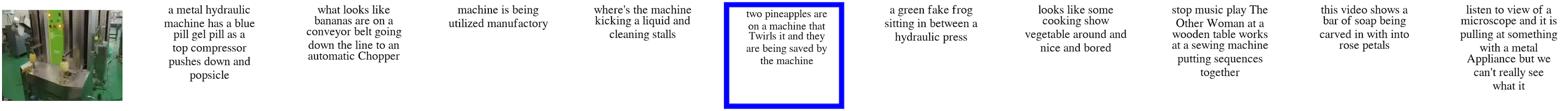}} \\ \hline
        (h) & \multicolumn{6}{c}{\includegraphics[width=1.0\linewidth, trim={0 0 41cm 0}, clip]{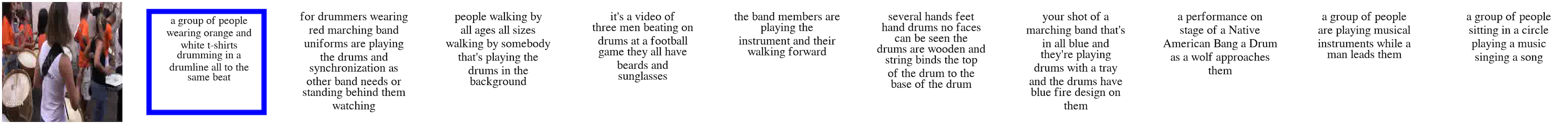}} \\ \hline
        (i) & \multicolumn{6}{c}{\includegraphics[width=1.0\linewidth, trim={0 0 41cm 0}, clip]{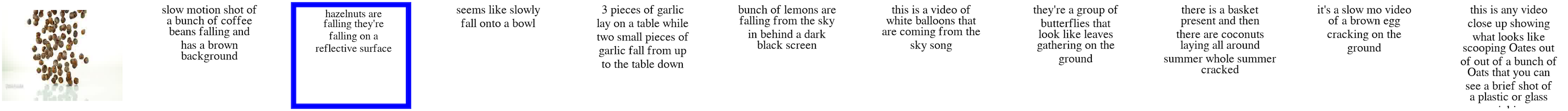}} \\
        \hline 
    \end{tabular}}
\caption{\textbf{Spoken Moments Examples of Video to Caption Retrieval Results:} Given a query video, we show five top retrieval captions where words transcribed from the audio captions using a pretrained ASR model are used as a caption. We use a BART model trained with the AMM loss function on the S-MiT dataset. Blue indicates the ground-truth results.}
\label{table:analysis_i2c}
\end{table*}

\section{Captions in the Spoken Moments Dataset}

Table~\ref{table:smit_example} shows some captions in the Spoken Moments dataset that capture motion and sequential events which would be difficult to represent with a single image.  
\begin{table*}[h]
\centering
\scalebox{0.8}{
    \begin{tabular}{ c | P{4cm} P{2.5cm} P{2.5cm} P{2.5cm} P{2.5cm} P{2.5cm}}
        \hline 
        
        \multirow{2}{*}{} & \multirow{2}{*}{\textbf{Caption}} & \multicolumn{5}{c}{\textbf{Frames}} \\
        & & \multicolumn{5}{c}{$\xrightarrow{\hspace*{13cm}}$ time} \\ 
        \hline 
        (a) & \footnotesize{a boy and a red white and blue shirt is sitting on a couch he is \textbf{holding} an infant life vest and \textbf{picks it up to blow} through the two}
        & \includegraphics[width=1.0\linewidth]{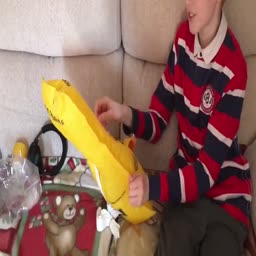}
        & \includegraphics[width=1.0\linewidth]{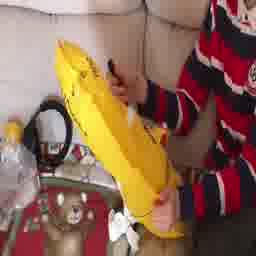}
        & \includegraphics[width=1.0\linewidth]{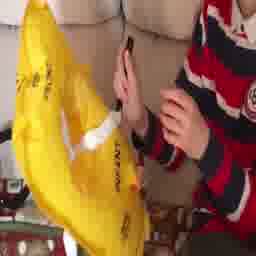}
        & \includegraphics[width=1.0\linewidth]{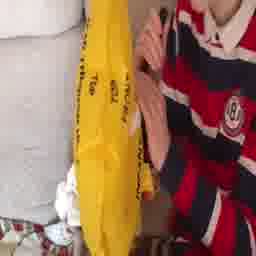}
        & \includegraphics[width=1.0\linewidth]{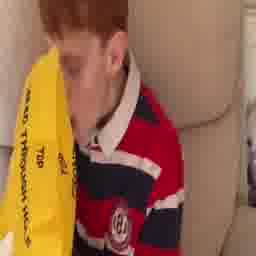}
        \\ \hline
        (b) & \footnotesize{there's a gauge or a lock thing turns from rides and then \textbf{being turned to the left}}
        & \includegraphics[width=1.0\linewidth]{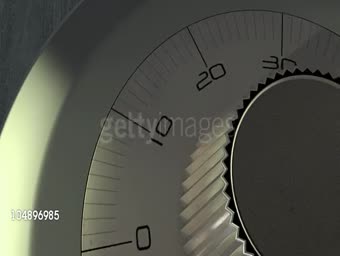}
        & \includegraphics[width=1.0\linewidth]{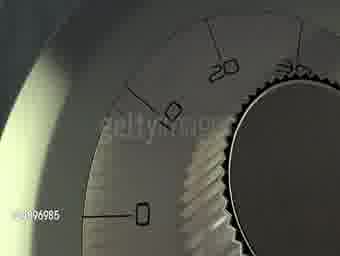}
        & \includegraphics[width=1.0\linewidth]{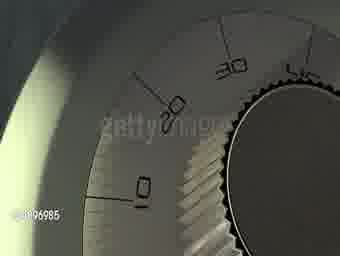}
        & \includegraphics[width=1.0\linewidth]{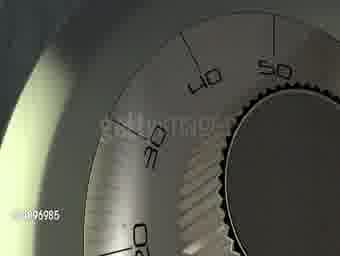}
        & \includegraphics[width=1.0\linewidth]{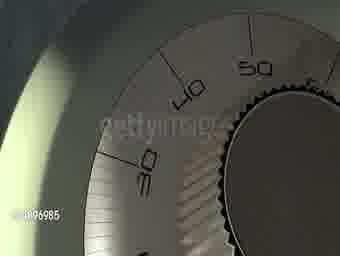}
        \\ \hline
        (c) & \footnotesize{a picture of a man \textbf{drinking coffee} and \textbf{play with a cell phone} in \textbf{fast motion}}
        & \includegraphics[width=1.0\linewidth]{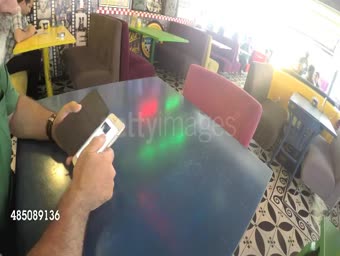}
        & \includegraphics[width=1.0\linewidth]{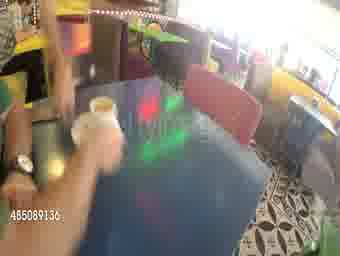}
        & \includegraphics[width=1.0\linewidth]{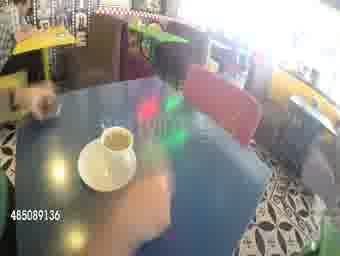}
        & \includegraphics[width=1.0\linewidth]{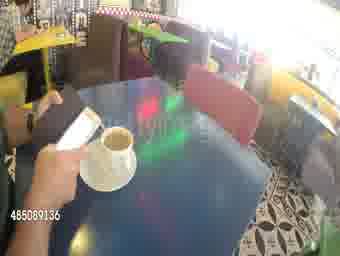}
        & \includegraphics[width=1.0\linewidth]{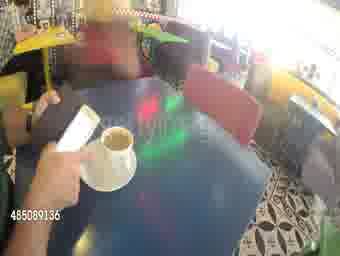}
        \\ \hline
        (d) & \footnotesize{in \textbf{slow motion} we see a collie \textbf{jump into the air} and \textbf{catch} a white frisbee in flight}
        & \includegraphics[width=1.0\linewidth]{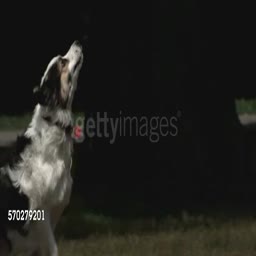}
        & \includegraphics[width=1.0\linewidth]{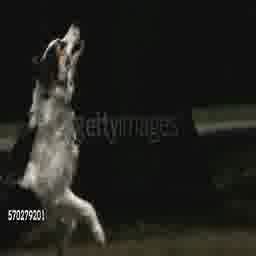}
        & \includegraphics[width=1.0\linewidth]{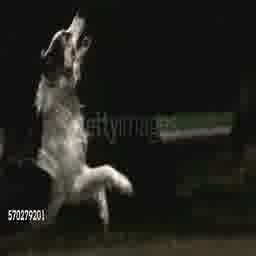}
        & \includegraphics[width=1.0\linewidth]{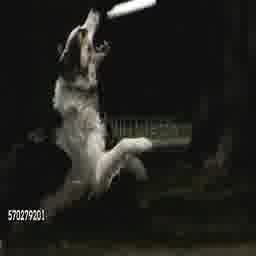}
        & \includegraphics[width=1.0\linewidth]{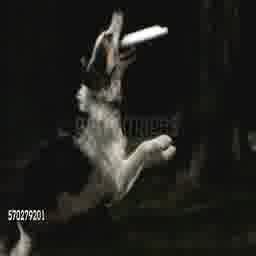}
        \\ \hline 
        (e) & \footnotesize{these are track and field runners and it's a relay race and \textbf{they take off when they are handed the batons}}
        & \includegraphics[width=1.0\linewidth]{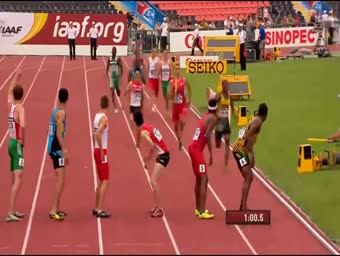}
        & \includegraphics[width=1.0\linewidth]{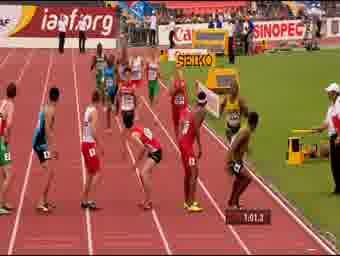}
        & \includegraphics[width=1.0\linewidth]{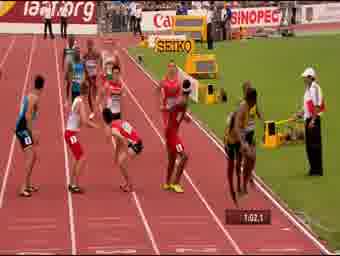}
        & \includegraphics[width=1.0\linewidth]{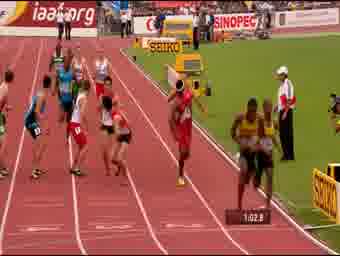}
        & \includegraphics[width=1.0\linewidth]{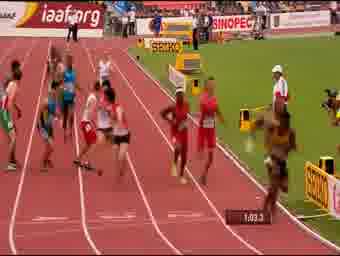}
        \\ \hline 
        (f) & \footnotesize{there is water \textbf{dripping off} the edge of something all you can hear is the water dripping}
        & \includegraphics[width=1.0\linewidth]{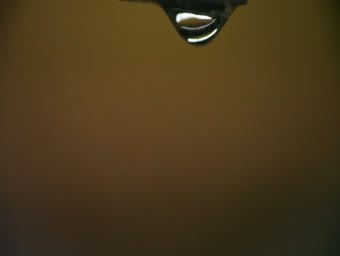}
        & \includegraphics[width=1.0\linewidth]{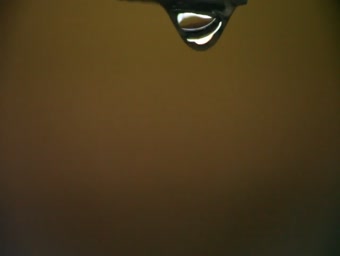}
        & \includegraphics[width=1.0\linewidth]{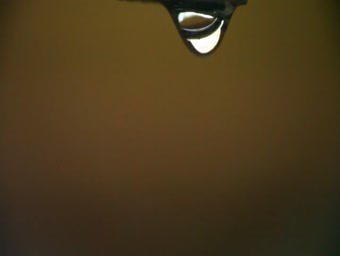}
        & \includegraphics[width=1.0\linewidth]{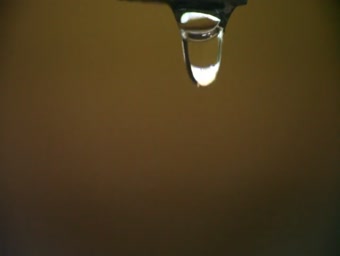}
        & \includegraphics[width=1.0\linewidth]{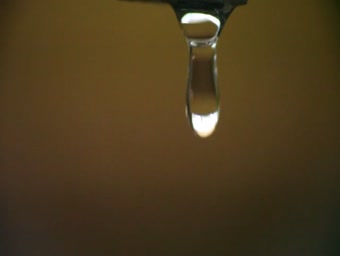}
        \\ \hline 
    \end{tabular}}
\caption{\textbf{Spoken Moments Captions:} We show some examples of captions, and associated video frames, from the Spoken Moments dataset, where the captions describe a sequence of actions or motion.}
\label{table:smit_example}
\end{table*}



\end{document}